\documentclass{article}
\usepackage{ijcai19}

\usepackage{times}
\usepackage{soul}
\usepackage{url}
\usepackage[hidelinks]{hyperref}
\usepackage[utf8]{inputenc}
\usepackage[small]{caption}
\usepackage{graphicx}
\usepackage{amsmath,amssymb,amsfonts,amsthm}
\usepackage{booktabs}
\usepackage{xcolor}

\newtheorem{theorem}{Theorem}

\newtheorem{proposition}{Proposition}
\newtheorem{definition}{Definition}

\definecolor{red}{rgb}{0,0,0}		

\title{Towards Robust ResNet: A Small Step but a Giant Leap}

\author{
Jingfeng Zhang$^1$\and
Bo Han$^2$\and
Laura Wynter$^3$\and
Bryan Kian Hsiang Low$^1$\And
Mohan Kankanhalli$^1$\\
\affiliations
$^1$Department of Computer Science, National University of Singapore\\
$^2$RIKEN Center for Advanced Intelligence Project\\
$^3$IBM Research, Singapore
\emails
{j-zhang@comp.nus.edu.sg, 
bo.han@riken.jp,
lwynter@sg.ibm.com, 
\{lowkh, mohan\}@comp.nus.edu.sg}
}

\begin{document}

\maketitle

\begin{abstract}
This paper presents a simple yet principled approach to boosting the robustness of the \emph{residual network} (ResNet) that is motivated by a dynamical systems perspective. Namely, a deep neural network can be interpreted using a partial differential equation, which naturally inspires us to characterize ResNet based on an explicit Euler method. 
This consequently allows us to exploit the step factor $h$ in the Euler method 
to control the robustness of ResNet in both its training and generalization. In particular, we prove that a small step factor $h$ can benefit its training and generalization robustness during backpropagation and 
forward propagation, respectively.
Empirical evaluation on real-world datasets corroborates our analytical findings that a small $h$ can indeed improve both its training and generalization robustness.
\end{abstract}

\section{Introduction}
\emph{Deep neural networks} (DNNs) have reached an unprecedented level of predictive accuracy in several real-world application domains (e.g., text processing~\cite{conneau2016very,zhang2015character}, image recognition~\cite{he2016deep,Huang2017DenselyCC}, and video analysis
due to their capability of approximating any universal function.
However, DNNs are often difficult to train due in a large part to the vanishing gradient phenomenon~\cite{glorot2010understanding,ioffe2015batch}. 
The \emph{residual network} (ResNet)~\cite{he2016deep} was proposed to alleviate this issue using its key component known as the skip connection, which creates a ``bypass'' path for information propagation~\cite{he2016identityv2}.

Nevertheless, it remains challenging to achieve robustness in training a very deep DNN.
As the DNN becomes deeper, it requires more careful tuning of its model hyperparameters (e.g., learning rate, number of layers, choice of optimizer) to perform well. This issue can be mitigated using normalization techniques such as \emph{batch normalization} (BN)~\cite{ioffe2015batch}, layer normalization~\cite{lei2016layer}, and group normalization~\cite{wu2018group}, among which BN is most widely used.
BN normalizes the inputs of each layer to enable robust training of DNNs. It has been shown that BN provides a smoothing effect of the optimization landscape~\cite{2018arXiv180511604S}, thus ameliorating the issue of a vanishing gradient~\cite{ioffe2015batch}.

Unfortunately, we have observed in our experiments (Section~\ref{section:judge_small_h_stabilizing_gradients_weights}) that a very deep DNN can potentially (and surprisingly) experience exploding gradients in shallow layers in early training iterations, even when it is coupled with BN.
As a result, its weights change drastically, which in turn causes violent feature transformations, 
hence impairing its robustness in training.
Our subsequent experiments in Appendix~\ref{Appendix:judge_small_h_stabilizing_gradients_weights} also show that the gradients are large and bumpy in deep layers in the later training iterations, which destabilizes the training \textcolor{red}{procedure}. 
Besides, noisy data (e.g., images with mosaics and texts with spelling errors) can adversely impact the training procedure, in which the noise may be amplified through forward propagation of features, thus degrading its robustness in generalization.

This paper presents a simple yet principled approach to boosting the robustness of ResNet in both training and generalization that is motivated by a dynamical systems perspective~\textcolor{red}{\cite{ChenNeuralODE,lu2017beyond,ruthotto2018deep,wennan_dynamic}}. Namely, 
a DNN can be interpreted using a partial differential equation, which naturally inspires us to characterize ResNet based on an explicit Euler method. 
This consequently allows us to exploit the step factor $h$ in the Euler method to control the robustness of ResNet in both its training and generalization.
In our work here, \emph{training robustness} refers to the stability of model training with an increasing depth, a larger learning rate, and different types of optimizer, while
\emph{generalization robustness} refers to how well a trained model generalizes to classify  \textcolor{red}{test data} whose distribution may not \textcolor{red}{match} that of the training data.
To analyze \textcolor{red}{the} effects of step factor $h$, we prove that a small $h$ can benefit the training and generalization robustness of ResNet during backpropagation and forward propagation, respectively.
Empirical evaluation on real-world vision- and text-based datasets corroborates our analytical findings that a small $h$ can indeed improve both the training and generalization robustness.
\section{Background and Notations}
Before delving into a robust ResNet (Section~\ref{section:Towards Robust ResNet}), we review the necessary background information which includes ResNet, batch normalization, and partial differential equations.
\\
\textbf{Brief description of ResNet.}
A ResNet is a variant of the DNN that exhibits competitive predictive accuracy and convergence properties~\cite{he2016deep}. 
The ResNet comprises many stacked residual blocks with a key component called the skip connection, each of which has the following structure:
\begin{equation}
 \mathbf{y}_n  \triangleq   \mathbf{x}_n + \mathcal{F} (\mathbf{x}_n)\ ,\quad
 \mathbf{x}_{n+1}  \triangleq  I(\mathbf{y}_n)
\label{booboo}
\end{equation}
for layer $n$ where $\mathbf{x}_n$ and $\mathbf{x}_{n+1}$ are, \textcolor{red}{respectively}, the input and output of residual block $n$, $\mathcal{F}$ is a residual block performing feature transformations (e.g., convolutional operations or affine transformation), and $I$ is a component-wise operation (e.g., ReLU function~\cite{Nair2010} or identity mapping~\cite{he2016identityv2}). 
Based on the core idea of the skip connection, variants of ResNet have been proposed for specific tasks such as DenseNet for image recognition~\cite{Huang2017DenselyCC} and VDCNN with shortcut connections for text classification~\cite{conneau2016very}.
\\
\textbf{Batch normalization (BN).}
BN has been widely adopted for training DNNs~\cite{ioffe2015batch} and normalizes the input of a layer over each mini-batch of the training data. Specifically, for the input $\mathbf{x} \triangleq (x^{(k)})^{\top}_{k=1,\ldots,d}$ of a layer with dimension $d$, BN first normalizes each scalar input feature $\hat{x}^{(k)} \triangleq (x^{(k)} - \mu^{(k)})/\sigma^{(k)}$ for $k=1,\ldots,d$ independently where $\mu\triangleq(\mu^{(k)}\triangleq\mathbb{E}[x^{(k)}])^{\top}_{k=1,\ldots,d}$ and $\sigma\triangleq(\sigma^{(k)}\triangleq\sqrt{\mathrm{Var}[x^{(k)}]})^{\top}_{k=1,\ldots,d}$ are computed over a mini-batch of size $m$. Then, it performs an affine transformation of each normalized scalar input feature $\hat{x}^{(k)}$ by $\textrm{BN}(x^{(k)}) \triangleq \gamma^{(k)}\hat{x}^{(k)}  + \beta ^{(k)}$ where $\gamma \triangleq (\gamma^{(k)})^{\top}_{k=1,\ldots,d}$ and $\beta \triangleq (\beta^{(k)})^{\top}_{k=1,\ldots,d}$ are learned during training.
BN has been shown to smooth the optimization landscape of DNNs~\cite{2018arXiv180511604S}. This offers more stable gradients for the robust training of DNNs, thus ameliorating the issue of a vanishing gradient~\cite{ioffe2015batch}.
\\
\textbf{Characterizing DNN with a partial differential equation (PDE).} 
A DNN performs nonlinear feature transformations of the input such that the transformed features can  match the corresponding target output (e.g., categorical labels for classification and continuous quantities for regression). To achieve this, a DNN transforms the input through multiple layers such that the transformed \textcolor{red}{features} in the last layer become linearly separable~\cite{haber2017stable}.

\begin{figure}
    \includegraphics[scale=0.06]{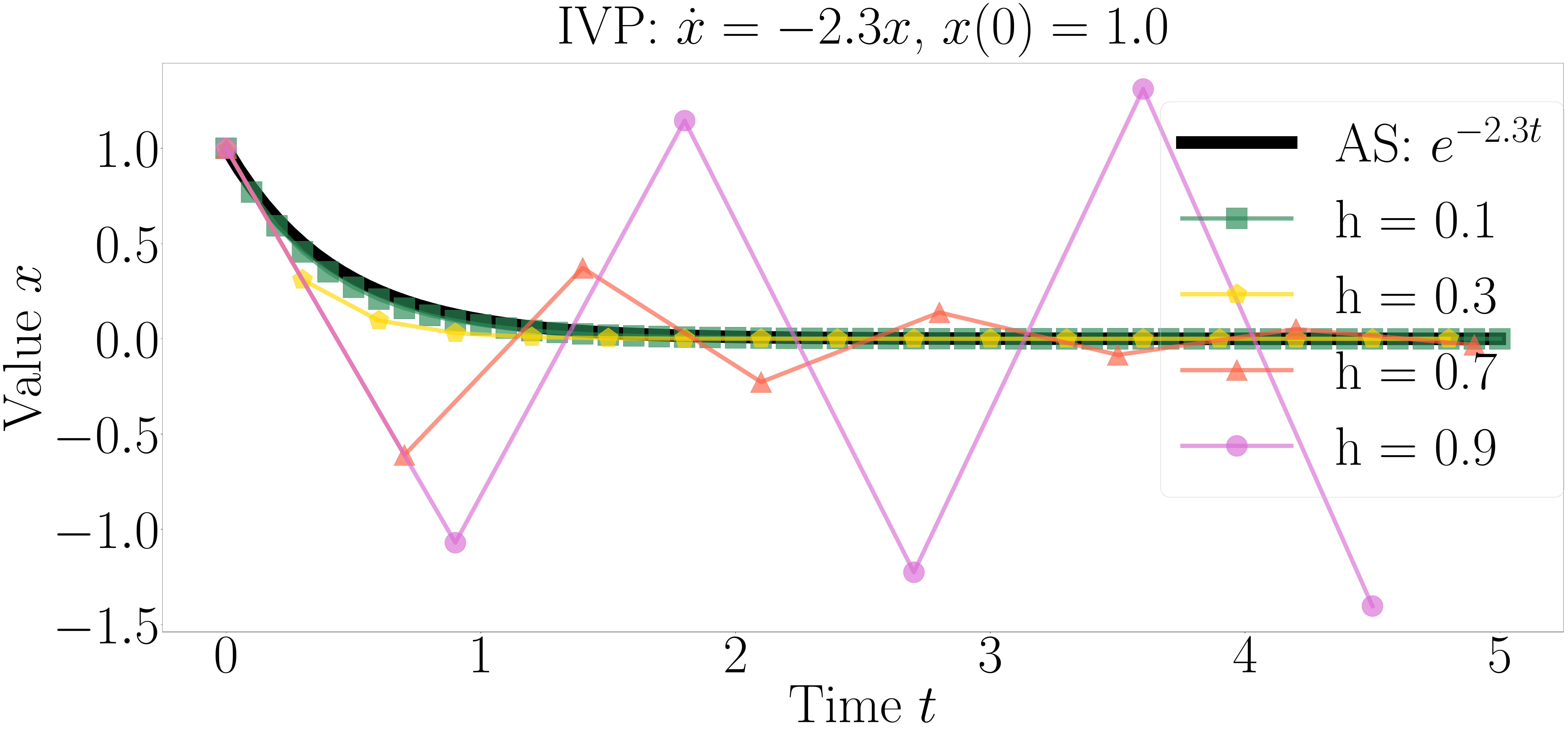}
  \caption{The \emph{analytic solution} (AS) of an IVP and its approximations by an explicit Euler method with different step factors $h$.}
  \label{fig:vis_euler_method}
\end{figure}

Let us now consider the dynamical systems perspective: A PDE can characterize the motion of particles~\cite{ascher2008numerical,atkinson2008introduction}.
This motivates us to characterize the feature transformations in a DNN as a system of first-order PDEs:
\begin{definition} The feature transformations in a DNN can be characterized by a first-order PDE $f: T \times X  \rightarrow \mathbb{R}^d$ where \textcolor{red}{$T \subseteq \mathbb{R}^+\cup \{0\}$} and $X \subseteq \mathbb{R}^d$:
	\begin{equation*}
	\dot{{\mathbf{x}}} \triangleq {\partial \mathbf{x}}/{\partial t} \triangleq f(t, \mathbf{x}) 
	\end{equation*}
	where $t \in T$ is the time along the feature transformations and
	$\mathbf{x} \in X $ is a feature vector of dimension $d$.
\end{definition}
\textcolor{red}{Let $\mathbf{x}(t)$ denote a transformed feature vector at time $t$.} 
Given an initial input feature vector $\mathbf{x}(0)$, a PDE $\dot{\mathbf{x}} = f(t, \mathbf{x})$ gives rise to the \emph{initial value problem} (IVP). 
A solution to an IVP is a function of time $t$. For example, consider the IVP: $\dot{x} = -2.3x$ and $x(0) = 1$. Then, its analytic solution can be derived as \textcolor{red}{$x(t) = \exp(-2.3 t)$}, as shown in Fig.~\ref{fig:vis_euler_method}. For more complicated PDEs, it is not always possible to derive an analytic solution. So, their solutions are approximated by numerical methods, of which the explicit Euler method (see Definition~\ref{Euler-Method} below) is the most classic example.  
\section{Towards Robust ResNet} 
\label{section:Towards Robust ResNet} 
%
A ResNet can naturally be described by the explicit Euler method (Definition~\ref{Euler-Method}), which gives us an Euler view of ResNet~\eqref{prol-Euler-viewed_resnet}. Inspired by the stability of the Euler method, 
the step factor $h$ for ResNet can enable 
smoother feature transformations  
due to gradual feature transitions at every consecutive block during forward propagation. The benefits of such smoothness are two-fold: (a) preventing information explosion over a large depth (Section~\ref{sec:training-robustness}) and (b) mitigating the adverse effect of noise in the input features (Section~\ref{sec:generalization-robustness}), both of which can significantly boost the robustness of ResNet. 
%
%
\subsection{Connections of ResNet with the Euler Method}
%
%
\begin{definition}\label{Euler-Method} The explicit Euler method~\cite{atkinson2008introduction,ascher2008numerical} can be represented by
	\begin{equation}
	\mathbf{x}_{n+1} \triangleq \mathbf{x}_{n} + h\ {f}(t_n, \mathbf{x}_n)
	\label{eq:h}
	\end{equation}
	where $\mathbf{x}_{0}\triangleq\mathbf{x}(0)$ and $\mathbf{x}_{n}$ is an approximation of $\mathbf{x}(t_n)$.
\end{definition}
Given the solution $\mathbf{x}(t_n)$ at time $t_n$, the PDE $\dot{\mathbf{x}} = f(t, \mathbf{x} )$ intuitively indicates ``in which direction to continue''. 
At time $t_n$, the explicit Euler method computes this direction $f(t_n, \mathbf{x}_n)$ and follows it over a small time step from $t_n$ to $t_n + h$. 
To obtain a reasonable approximation, the step factor $h$ in the \textcolor{red}{Euler method} has to be chosen to be ``sufficiently small''. Its magnitude depends on the PDE and its given initial input. For instance, in Fig.~\ref{fig:vis_euler_method}, we use the Euler method with various step factors $h$ to approximate \textcolor{red}{the solution of} the IVP: $\dot{x} = -2.3x$ and $x(0) = 1$.
It can be observed that the approximation of $x$ improves with \textcolor{red}{a smaller} $h$.\\
\textbf{Euler view of ResNet.} By setting $h = 1$ and ${f}(t_n, \mathbf{x}_n) =\mathcal{F} (\mathbf{x}_n )$, it can be observed that the Euler method~\eqref{eq:h} characterizes the forward propagation of the original ResNet structure~\eqref{booboo}~\cite{haber2017stable}. 
To generalize ResNet using the Euler method~\eqref{eq:h}, we characterize the forward propagation of an Euler view of the ResNet in the following way:
%
Following the convention in~\cite{he2016identityv2}, 
ResNet stacks multiple residual blocks with step factor $h \in \mathbb{R}^+$,
each of which has the following structure:
%
\begin{equation} 
\mathbf{y}_n \triangleq \mathbf{x}_n + h\ \mathcal{F} (\mathbf{x}_n, W_n, \textrm{BN}_n)\ , \quad \mathbf{x}_{n+1} \triangleq I(\mathbf{y}_n)
\label{prol-Euler-viewed_resnet}
\end{equation}
%
for layer $n$ where
batch normalization $\textrm{BN}_n$ applies directly after (or before) the feature transformation function parameterized by weights $W_n$ (e.g., convolutional operations or affine transformation matrix). 

For any fixed $t$, the Euler method provides a more refined approximation with a smaller $h$, as analyzed in~\cite{butcher2016numerical}. 
Can a small $h$ also improve the robustness of a deep ResNet in training and generalization?
To answer this question, we will prove that a small $h$ can prevent information explosion during backpropagation over a large depth (\textcolor{red}{Section~\ref{sec:training-robustness}}) and mitigate  the adverse effect of noise in the input features during forward propagation  (Section~\ref{sec:generalization-robustness}). 
%
%
%
\subsection{Training Robustness: Effect of Small $h$ on Information Backpropagation}\label{sec:training-robustness}
To simplify the analysis, let $I$ in~\eqref{prol-Euler-viewed_resnet} be an identity mapping, that is, $\mathbf{x}_{n+1} = \mathbf{y}_n$. Then, by recursively applying~\eqref{prol-Euler-viewed_resnet}, 
\begin{equation}
\mathbf{x}_N = \mathbf{x}_n + h \sum_{i=n}^{N-1} \mathcal{F}(\mathbf{x}_i, W_i, \textcolor{red}{\textrm{BN}}_i)
\label{recursive_resblock} 
\end{equation}
for any block $N$ deeper than any shallower block $n$.
We will use~\eqref{recursive_resblock} to analyze the information backpropagation. Let the loss function be denoted by $L$. Using chain rule, 
\begin{equation}
\frac{\partial L}{\partial \mathbf{x}_n}\hspace{-0.5mm} =\hspace{-0.5mm}
\frac{\partial L}{\partial \mathbf{x}_N} \frac{\partial \mathbf{x}_N}{\partial \mathbf{x}_n} \hspace{-0.5mm}=\hspace{-0.5mm}
\frac{\partial L}{\partial \mathbf{x}_N}\hspace{-1mm} \left(\hspace{-0.5mm}\mathbf{1}\hspace{-0.5mm}+\hspace{-0.5mm} h \frac{\partial}{\partial \mathbf{x}_n}\hspace{-0.5mm} \sum_{i=n}^{\textcolor{red}{N-1}} \hspace{-0.5mm}\mathcal{F}(\mathbf{x}_i, W_i, \textrm{BN}_i)\hspace{-0.5mm} \right)
\label{back_prop}
\end{equation}
where $\mathbf{1}$ denotes an identity matrix.
From~\eqref{back_prop}, the backpropagated information ${\partial L}/{\partial \mathbf{x}_n}$ can be decomposed into two additive terms: (a) The first term ${\partial L}/{\partial \mathbf{x}_N}$ propagates information directly without going through the weights of any layer, and (b) the second term $h ({\partial L}/{\partial \mathbf{x}_N}) {\partial}(\sum_{i=n}^{N-1} \mathcal{F} (\mathbf{x}_i, W_i, \textrm{BN}_i) )/{\partial \mathbf{x}_n}$ propagates through the weights of layers $n,\ldots,N-1$.
The first term ensures that information ${\partial L}/{\partial \mathbf{x}_n}$ does not vanish~\cite{he2016identityv2}. 
However, the second term can blow up the information ${\partial L}/{\partial \mathbf{x}_n}$, especially when the weights of layers $n,\ldots,N-1$ are large. A standard approach to resolving this issue is to apply BN after the \textcolor{red}{feature transformation function}~\cite{ioffe2015batch}.

Let us first study the effect of BN. To ease analysis, we examine the residual block $n$ where $N = n+1$, $\mathcal{F} (\mathbf{x}_n, W_n, \textrm{BN}_n) \triangleq \textrm{BN}_n(\mathbf{x}'_n \triangleq W_n \mathbf{x}_n)$, 
and $\textrm{BN}_n(\mathbf{x}'_n) \triangleq \gamma ({\mathbf{x}'_n - {\mu}})/{{\sigma}} + \beta$ is a component-wise operation. 
Let $\sigma_n$ denote the smallest component in $\sigma$. Then, 
$\lVert{\partial \mathcal{F}}(\mathbf{x}_n)/{\partial \mathbf{x}_n} \rVert \leq \lVert\gamma\rVert\lVert W_n\rVert/{\sigma_n}$. Consequently, it follows from~\eqref{back_prop} that 
\begin{equation}
\left\lVert \frac{\partial L}{\partial \mathbf{x}_n} \right\rVert \leq
\left\lVert \frac{\partial L}{\partial \mathbf{x}_{n+1}} \right\rVert \left(1+ \frac{h}{\sigma_n} \lVert\gamma\rVert \lVert W_n \rVert \right). 
\label{one_block_judge_depth}
\end{equation} 
As observed in~\cite{2018arXiv180511604S}, $\sigma$ tends to be large. So, BN has the effect of constraining the explosive backpropagated information~\eqref{one_block_judge_depth}. However, as a ResNet grows deeper, $\sigma$ tends to be highly uncontrollable in practice and the backpropagated information still accumulates over a large depth and can once again blow up. 
For example, Fig.~\ref{vis:judge_gradients_and_weights} shows that its gradients are very large and unstable in early training iterations (red line). As a result, when a deep ResNet is trained on the CIFAR-10 or AG-NEWS dataset, its performance is much worse than that of its shallower counterparts, as shown in Fig.~\ref{fig:judge_depth} (red line).
In contrast, Fig.~\ref{vis:judge_gradients_and_weights} also shows that a reduced $h$ can serve to re-constrain the explosive backpropagated information (blue line). 
In fact, even without BN, reducing $h$ can still serve to stabilize the training procedure.

Our first theoretical result analyzes the effect of small $h$ on information backpropagation: 
\begin{proposition}
\label{yesyesyes}
	\textcolor{red}{Let} $n=0$ and $N = D$ where $D$ is the index of last residual block.
Suppose that $\lVert{\partial \mathcal{F}}(\mathbf{x}_i)/{\partial \mathbf{x}_i} \rVert \leq \lVert\gamma\rVert\lVert W_i\rVert/{\sigma_i}\leq \mathcal{W}$ for $i=0,\ldots, D-1$.
%
	Then, 
\begin{equation*}	
	\left\lVert\frac{\partial L}{\partial \mathbf{x}_0}\right\rVert \leq \left\lVert\frac{\partial L}{\partial \mathbf{x}_D}\right\rVert \left(\sqrt{d}+h\mathcal{W}\right)^D .
\end{equation*}	 
\end{proposition}
Its proof is in Appendix~\ref{Appendix: Proof_of_Prop_1}.
Note that $\mathcal{W}$ intuitively upper bounds the effects of BN. Also, since $\sqrt{d}+h\mathcal{W}\geq 1$,  the backpropagated information explodes exponentially w.r.t. the depth $D$, which directly affects the gradients and hurts the training robustness of a ResNet. Fortunately, reducing $h$ can give extra control of the backpropagated information since the term $(\sqrt{d}+h\mathcal{W})^D$ can be constrained by $h$ and the backpropagated information ${\partial L}/{\partial \mathbf{x}_0}$ is thus less likely to explode. This demonstrates that when a ResNet grows deeper, the step factor $h$ should be reduced to a smaller value.
\subsection{Generalization Robustness: Effect of Small $h$ on Information Forward Propagation}\label{sec:generalization-robustness}
It can be observed from~\eqref{prol-Euler-viewed_resnet} that a reduced $h$ gives extra control of the feature transformations in a ResNet by making them smoother, that is, smaller $\lVert\mathbf{x}_{n+1} -\mathbf{x}_{n} \rVert$.
More importantly, as the features propagate forward through a deep ResNet, the adverse effect of the  noise in the input features can be mitigated over a large depth. In particular, we will prove that a reduced $h$ can help to stabilize the target output  of a ResNet against noise in the input features.

Let $\mathbf{x}^{\epsilon}_0$ be a perturbation from $\mathbf{x}_0$ (i.e., $\lVert\mathbf{x}^{\epsilon}_0 - \mathbf{x}_0\rVert \leq \epsilon$) and $\mathbf{x}^{\epsilon}_{N}$ be the corresponding transformed feature vector in the ResNet. Then, from~\eqref{recursive_resblock},
%
\begin{equation} 
\begin{array}{rl}
\lVert\mathbf{x}^{\epsilon}_{N} - \mathbf{x}_N \rVert =\hspace{-2.4mm}&\displaystyle
\left\lVert \mathbf{x}^{\epsilon}_0 + h \sum_{i=0}^{N-1}\mathcal{F}(\mathbf{x}^{\epsilon}_i)  -  \mathbf{x}_0 - h \sum_{i=0}^{N-1} \mathcal{F}(\mathbf{x}_i) \right\rVert \\ 
\leq \hspace{-2.4mm}&\displaystyle
\epsilon + h  \sum_{i=0}^{N-1} \left\lVert \mathcal{F}(\mathbf{x}^{\epsilon}_i)  - \mathcal{F}(\mathbf{x}_i)\right\rVert .
\end{array}
\label{theoretical_judge_noise}
\end{equation}
It can be observed from~\eqref{theoretical_judge_noise} that the noise in the input features is amplified with an increasing depth of the ResNet. Fortunately, by introducing a reduced $h$, the noise amplification can be limited and its adverse effect can thus be mitigated.

Our next theoretical result analyzes the effect of small $h$ on information forward propagation:
\begin{proposition}
	Consider the last residual block \textcolor{red}{$N = D$}. Let the noise in the last layer $D$ be denoted by $\epsilon_{D}\triangleq \lVert\mathbf{x}^{\epsilon}_{D} - \mathbf{x}_D\rVert$.
	Suppose that $\lVert\mathcal{F}(\mathbf{x}^{\epsilon}_i)  - \mathcal{F}(\mathbf{x}_i)\rVert \leq \mathcal{W}$ for $i=0,\ldots,D-1$.
	 Then, 
	\begin{equation}
	\epsilon_{D} \leq \epsilon + hD \mathcal{W}\ .
	  \label{eliminate_noise}
	\end{equation}
\end{proposition}
Its proof follows directly from~\eqref{theoretical_judge_noise}. 
It can be observed from~\eqref{eliminate_noise} that a small $h$ can mitigate the adverse effect of noise that is accumulated over a large depth $D$. Hence, a deeper ResNet (i.e., larger $D$) requires a smaller $h$ while a shallower ResNet allows for a larger $h$.

Note that for a given depth of ResNet, the step factor $h$ cannot be reduced to be infinitesimally small. In the limiting case of $h = 0$, though the noise after the feature transformations would be perfectly bounded, there is no transformation of the initial input feature vector $\mathbf{x}_0$, that is, all feature transformations are smoothed out. 
%
\section{Experiments and Discussion}
In this section, we conduct experiments on the vision-based CIFAR-10 dataset~\cite{krizhevsky2009learning} and the text-based AG-NEWS dataset~\cite{zhang2015character}.
We also employ a synthetic binary TWO-MOON dataset in Section~\ref{exp:generalization-robustness} to illustrate how the adverse effect of noise in the input features is mitigated along the forward propagation. We fix our step factor $h=0.1$ and compare it with the original ResNet (i.e., $h=1$) in Sections~\ref{exp:training-robust} and~\ref{exp:generalization-robustness}. We will discuss how to select the step factor $h$ in Section~\ref{exp:selection-of-h}.

For the vision-based CIFAR-10 dataset, the residual block $\mathcal{F}$ contains two $2$D convolutional operations~\cite{he2016deep}. For the text-based AG-NEWS dataset, the residual block $\mathcal{F}$ contains two $1$D convolutional operations~\cite{conneau2016very}. For the synthetic TWO-MOON dataset, the residual block $\mathcal{F}$ contains two affine transformation matrices. 

To tackle the dimensional mismatch in the shortcut connections of ResNet, we adopt the same practice as that in~\cite{he2016deep} for the CIFAR-10 dataset and in~\cite{conneau2016very} for the AG-NEWS dataset by using convolutional layers with the kernel of size one to match the dimensions.
Unless specified otherwise, the default optimizer is SGD with $0.9$ momentum. We train a ResNet using the  CIFAR-10 dataset for $80$ epochs with an initial \emph{learning rate} (LR) of $0.1$ that is divided by $10$ at epochs $40$ and $60$. We train another ResNet using the AG-NEWS dataset with a fixed LR of $0.1$ for $15$ epochs. 
\subsection{Small $h$ Improves Training Robustness}\label{exp:training-robust}
\begin{figure} 
\begin{tabular}{c}
    \includegraphics[scale=0.167]{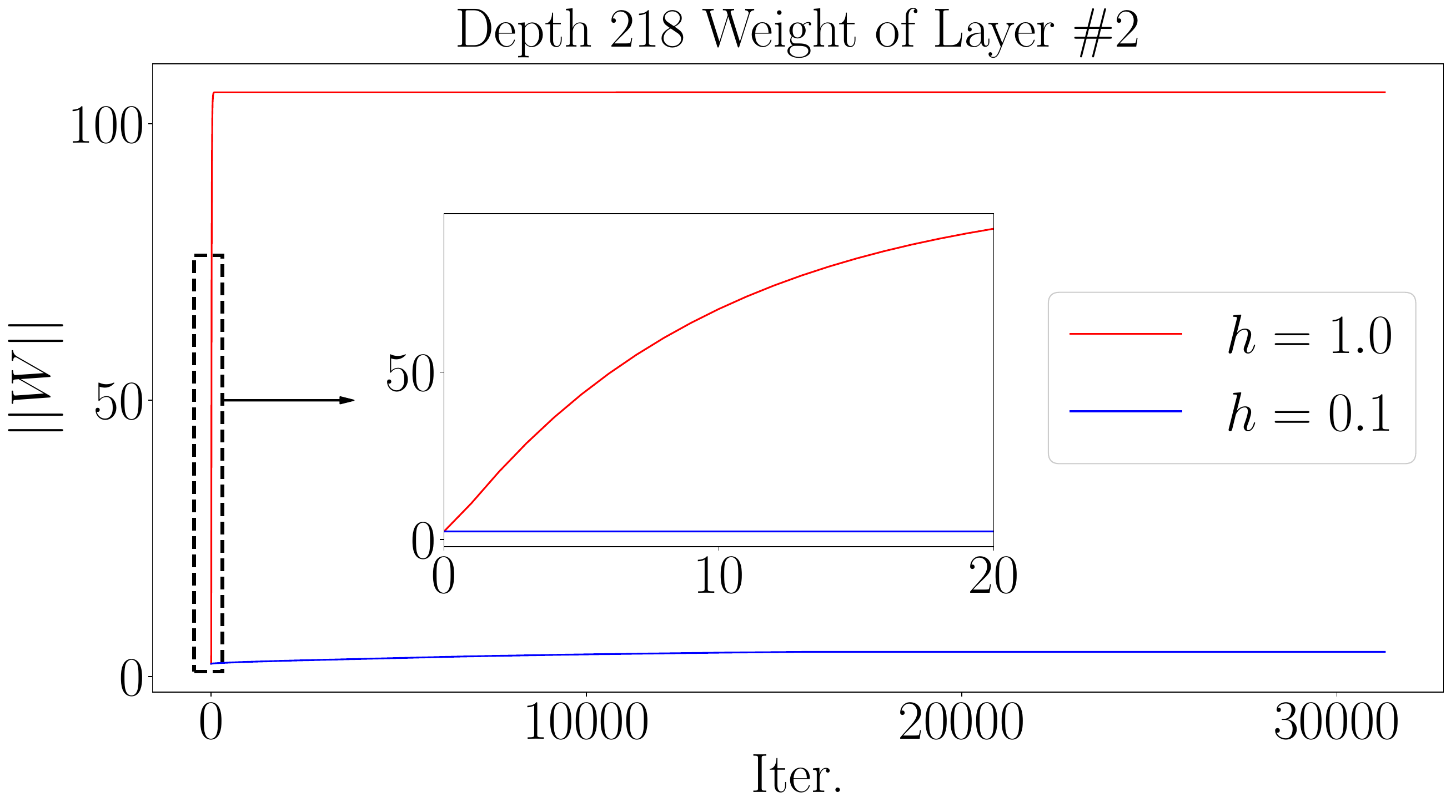}\\
    \includegraphics[scale=0.167]{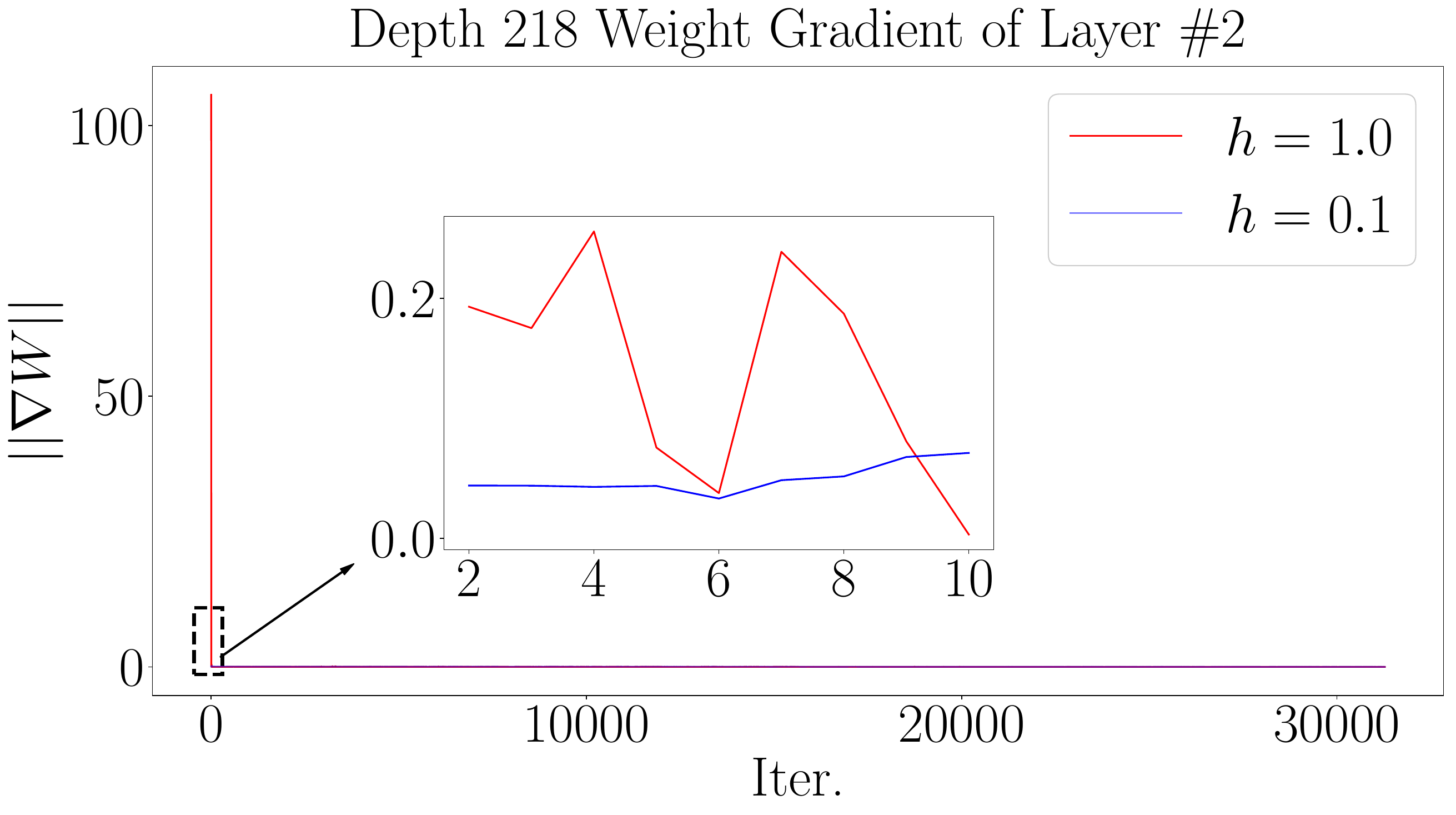}
\end{tabular}    
	\caption{Dynamics of weights (top) and corresponding gradients (bottom) of layer $2$ of ResNet with depth $218$ over training iterations.}
	\label{vis:judge_gradients_and_weights}
\end{figure}

\paragraph{Small $h$ stabilizes gradients and encourages smaller weights.}
\label{section:judge_small_h_stabilizing_gradients_weights}
To illustrate why a small $h$ can give extra training robustness as compared with the original ResNet (i.e., $h = 1.0$), we train a ResNet with depth $218$ on the CIFAR-10 dataset and collect statistics regarding the weights and corresponding gradients of layer $2$, as shown in Fig.~\ref{vis:judge_gradients_and_weights}.  Such statistics for ResNets with depths $110$ and $218$ over different layers are detailed in Appendix~\ref{Appendix:judge_small_h_stabilizing_gradients_weights}. 
\\
\textbf{Analysis 1 (Exploding gradients).} Fig.~\ref{vis:judge_gradients_and_weights} and the other figures in Appendix~\ref{Appendix:judge_small_h_stabilizing_gradients_weights} show that in the early training iterations, for the case of large $h$ (i.e., $h=1$), there are exploding gradients which make the weights quickly reach a plateau in a few initial iterations (red line).
In the early training iterations, this causes violent feature transformations, which may amplify the noise in the training inputs, thus degrading the performance of ResNet. 
Furthermore, as shown in Appendix~\ref{Appendix:judge_small_h_stabilizing_gradients_weights}, in the later training iterations, a ResNet with a larger $h$ tends to have larger and bumpy gradients in its deep layers (red line); this may cause the issue of overshooting, thus adversely affecting its convergence. 

The philosophy of characterizing a ResNet using a PDE is that the features should transform gradually over a large depth such that the transformed features in the last layer are linearly separable. So, we do not favor unstable gradients that cause the weights to change drastically, which in turn causes violent feature transformations. 
\\
\textbf{Analysis 2 (Large weights).} It can also be observed from Fig.~\ref{vis:judge_gradients_and_weights} that a larger $h$ (e.g., $h = 1$) tends to encourage larger weights during the training procedure (red line). To show this, we calculate the averaged weights of layer $2$ across all iterations. We repeat these experiments over two different random seeds. The case of $h = 1$ yields averaged weights of $51.12$ and $105.63$ while the case of $h = 0.1$ yields much smaller averaged weights of $4.54$ and $4.10$. 

As shown in~\cite{ketkar2017deep}, a trained model with large weights is more complex than that with small weights and tends to indicate overfitting to the training data. In practice, it is preferable to choose  simpler models (i.e., Occam's razor principle). So, we favor models with smaller weights. 
In addition, as stated in~\cite{reed1999neural}, large weights make the network unstable such that minor variation or statistical noise in the inputs will result in large differences in the target output. 

\begin{figure}
\centering
	\includegraphics[scale=0.24]{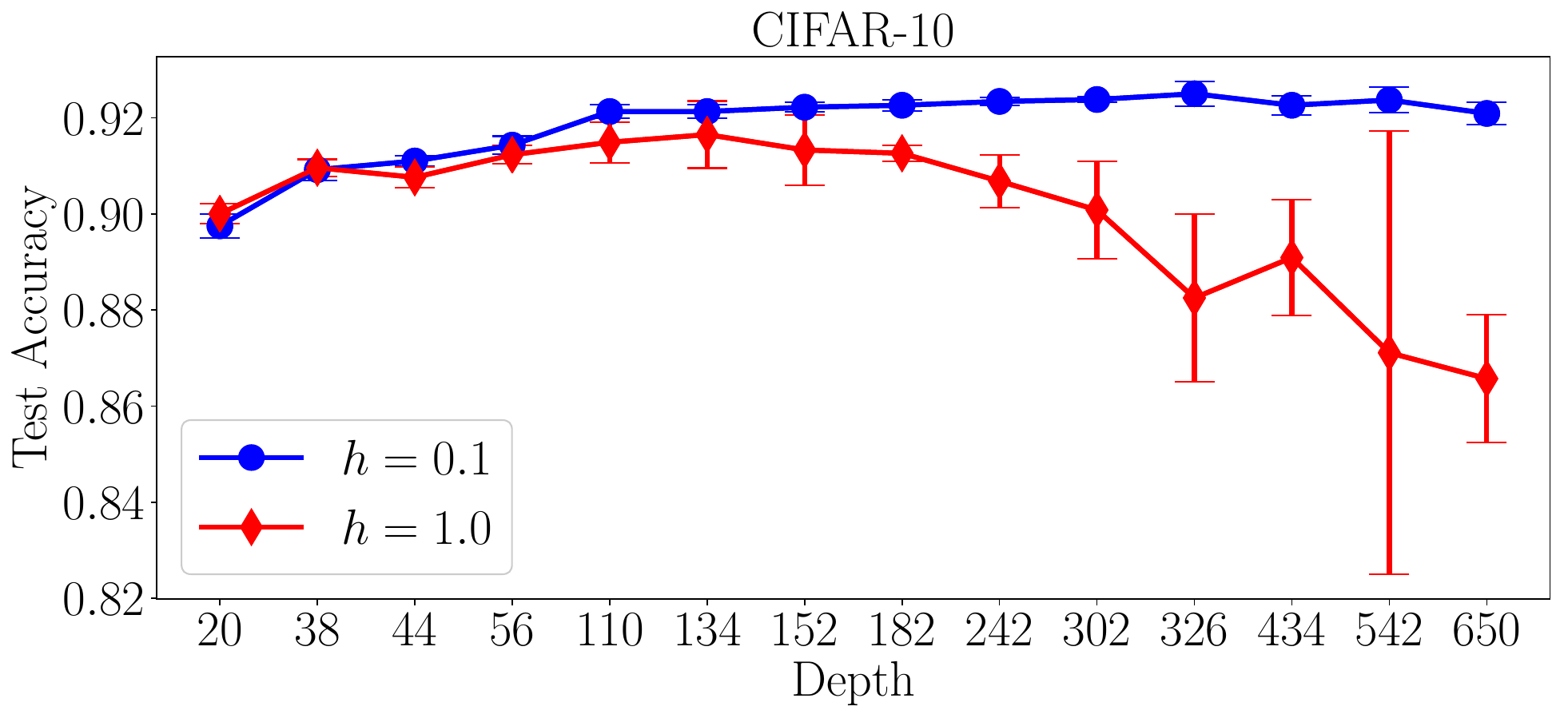} 
	\includegraphics[scale=0.24]{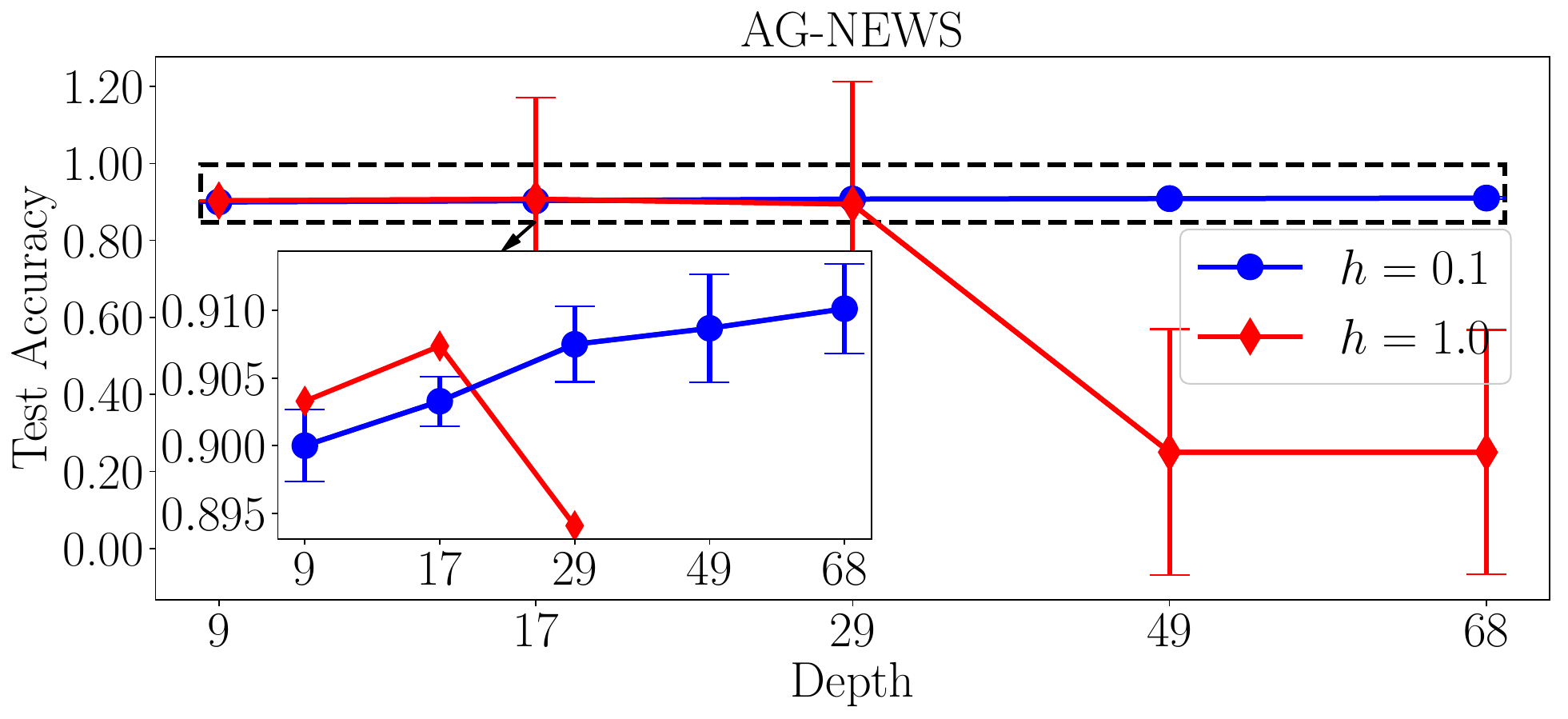}
	\caption{Comparison of training robustness of ResNets with increasing depths.}
	\label{fig:judge_depth}
\end{figure}

To mitigate such an issue, a number of techniques have been proposed: (a) Warming up the training with a small LR~\cite{he2016deep} and gradient clipping can counter the side effects of the issues of large and bumpy gradients, and 
(b) weight decay (e.g., L$2$ regularizer) can provide an extra penalty on large weights during training  such that it encourages using a simpler DNN represented by smaller weights. 
However, these techniques require tedious fine-tuning by highly trained domain experts 
(e.g., how many iterations to run in the warm-up phase of training, how to set LR to be sufficiently small, how much to penalize weights during training). 
Furthermore, adding weight decay introduces extra computation to train a DNN and choosing a smaller LR will significantly affect training convergence, which requires more computational power for convergence. 
In contrast, without employing specialized techniques, our proposed use of a smaller $h$ (e.g., $h = 0.1$) can serve to stabilize gradients over all training iterations; it encourages smaller weights for a deep ResNet, as shown in Fig.~\ref{vis:judge_gradients_and_weights} (blue line).
In addition, a small $h$ does not introduce extra computation and is compatible with the above techniques.
\paragraph{Small $h$ enables training deeper ResNets.}
To verify the training robustness of a small $h$ with an increasing depth of ResNet,  Fig.~\ref{fig:judge_depth} compares a ResNet with a reduced step factor $h = 0.1$ with the original ResNet (i.e., $h = 1$) over varying depths. Each configuration has $5$ trials with different random seeds.  We provide the median test accuracy with the standard deviation plotted as the error bar.

In both CIFAR-10 and AG-NEWS datasets, our ResNet with $h = 0.1$ outperforms that with $h = 1$ when their depths increase. In particular, \textcolor{red}{as} the depth of the ResNet increases, the original ResNet with $h=1.0$ experiences a performance degradation while our ResNet with $h = 0.1$ has an increasing or stabilizing performance.
It is also be observed that the blue shaded area (i.e., $h = 0.1$) is much thinner than the red one (i.e., $h = 1$). This shows that our ResNet with $h = 0.1$ has a smaller variance of the test accuracy over different random seeds.
This demonstrates that a small $h$ offers training robustness as well.

To summarize, as proven in Section~\ref{sec:training-robustness}, a reduced $h$ can prevent the explosion of backpropagated information over a large depth, thus making the training procedure of ResNet more robust to an increasing depth.

In \textcolor{red}{terms} of training robustness, a ResNet with a small $h$ is also robust to larger learning rates and different types of optimizer. In fact, even without BN, a \textcolor{red}{small} $h$ still helps to stabilize the training procedure while the performance of the original ResNet (i.e., $h = 1.0$) degrades significantly. 

\paragraph{Small $h$ enables large learning rate (LR).}
\label{section:judge_small_h_enable_larger_LR}
\begin{figure} \centering

	\includegraphics[scale=0.25]{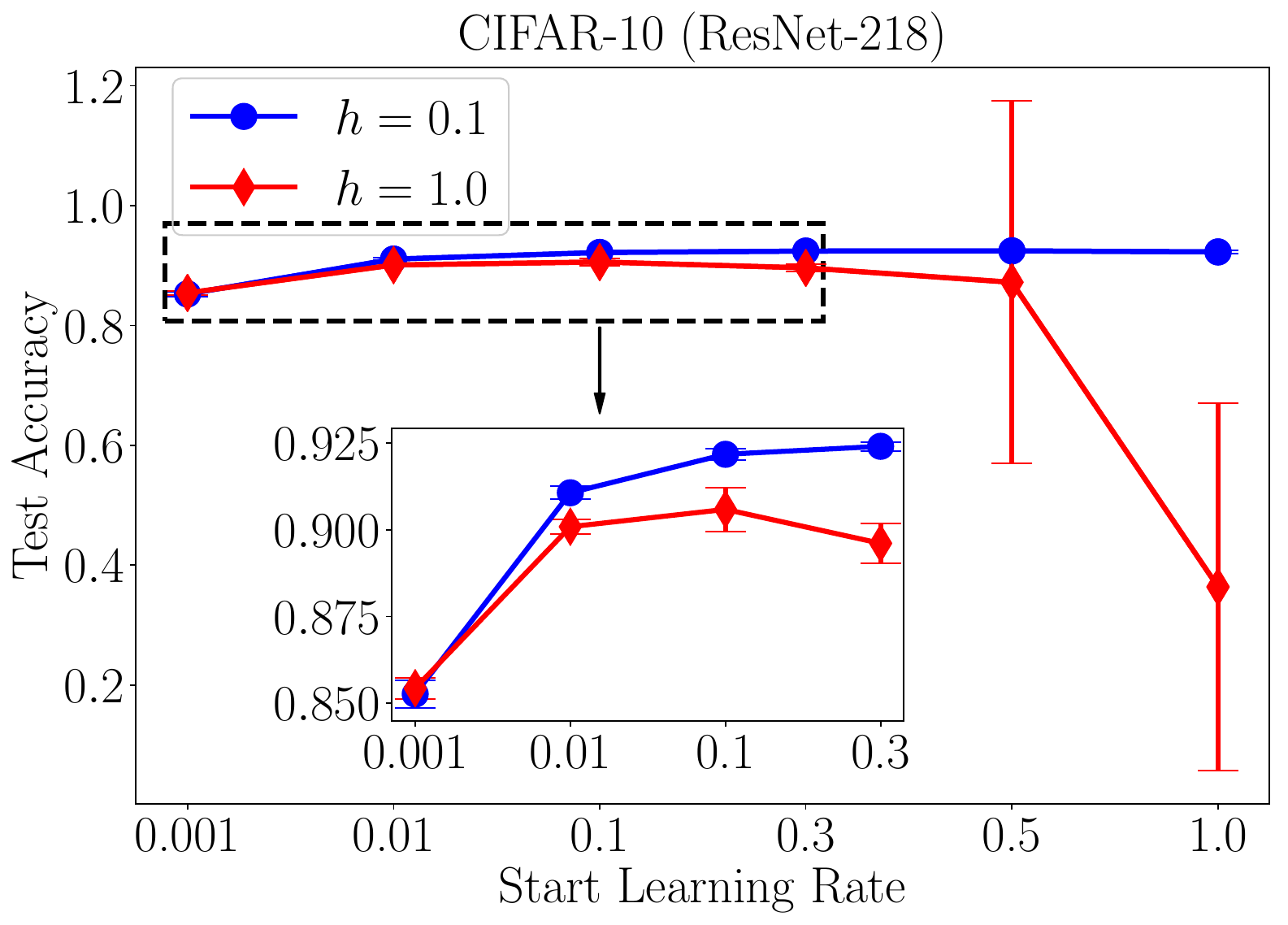}

	\caption{Training robustness comparisons of ResNet with different learning rates.}
	\label{fig:vis_different_leanring_rate}
\end{figure}
In Fig.~\ref{fig:vis_different_leanring_rate}, we train ResNet with depth 218 for CIFAR-10 dataset and compare its training performance over different learning rates.
Each configuration has $5$ trials with different random seeds.  We provide the median test accuracy with the standard deviation plotted as the error bar.
In addition, we put more results in Appendix~\ref{Appendix:judge_small_h_enable_larger_LR}, i.e, ResNet with depth 29 and 49 for {AG-NEWS}, and ResNet with depth 44 and 110 for {CIFAR-10}. 
We also illustrate the convergence rate with different LR of test accuracy over epochs in Appendix~\ref{Appendix:judge_small_h_enable_larger_LR}.

Fig.~\ref{fig:vis_different_leanring_rate} shows that our method, i.e., smaller $h$ (blue line), can enable larger LR for training without degrading its performance, due to that stable and smaller gradient throughout all training iterations over all layers (analyzed above). 
Compared with the larger $h$ (red line), our method also has a narrow error bar, which signifies smaller performance variance. 
In addition, as shown in the Figure~\ref{fig:vis_different_leanring_rate_convergence} in Appendix~\ref{Appendix:judge_small_h_enable_larger_LR}, our method (blue line) has faster convergence rate by utilizing larger training rate, e.g, start LR $= 0.3$. 
To sum up, small $h$ makes the training procedure of ResNet more robust to larger LR so that it can enjoy the benefits using larger LR, i.e., faster convergence, with little chance of performance degradation. 

\paragraph{Small $h$ helps networks without applying BN.}
\label{section:judge_small_h_without_BN}
To verify the effectiveness of small $h$ on improving robustness of the training procedure without applying BN, we compare ResNet with reduced step factor $h$ ($h=0.1$) and the original ResNet (corresponding to $h = 1.0$) without BN (other hyperparameters remain the same). 
Each configuration has 5 trials with different random seeds. We provide median test accuracy with the standard deviation plotted as the error bar.
\begin{figure}[tp!]
	\centering
	\includegraphics[scale=0.25]{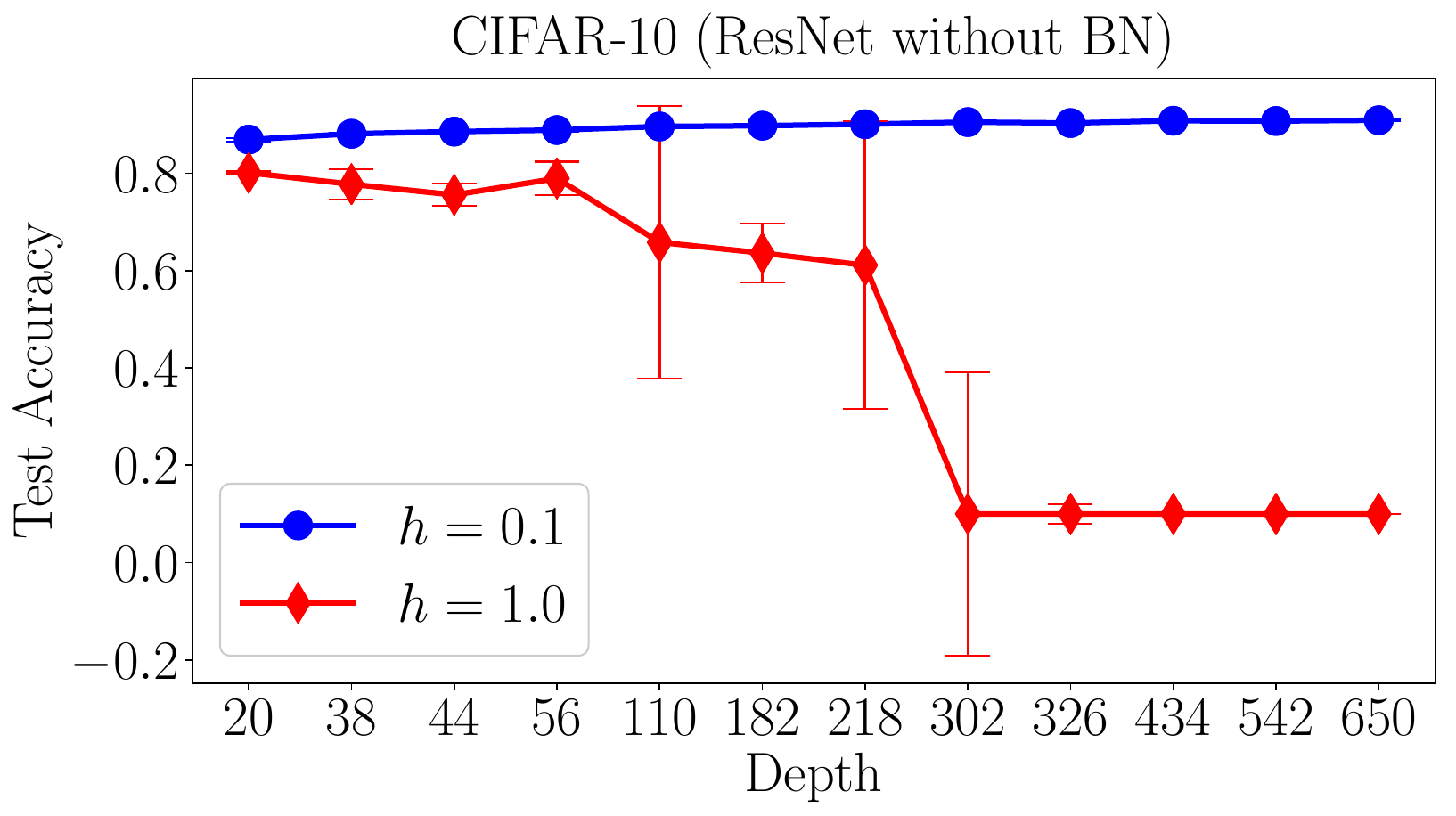}
	\includegraphics[scale =0.25]{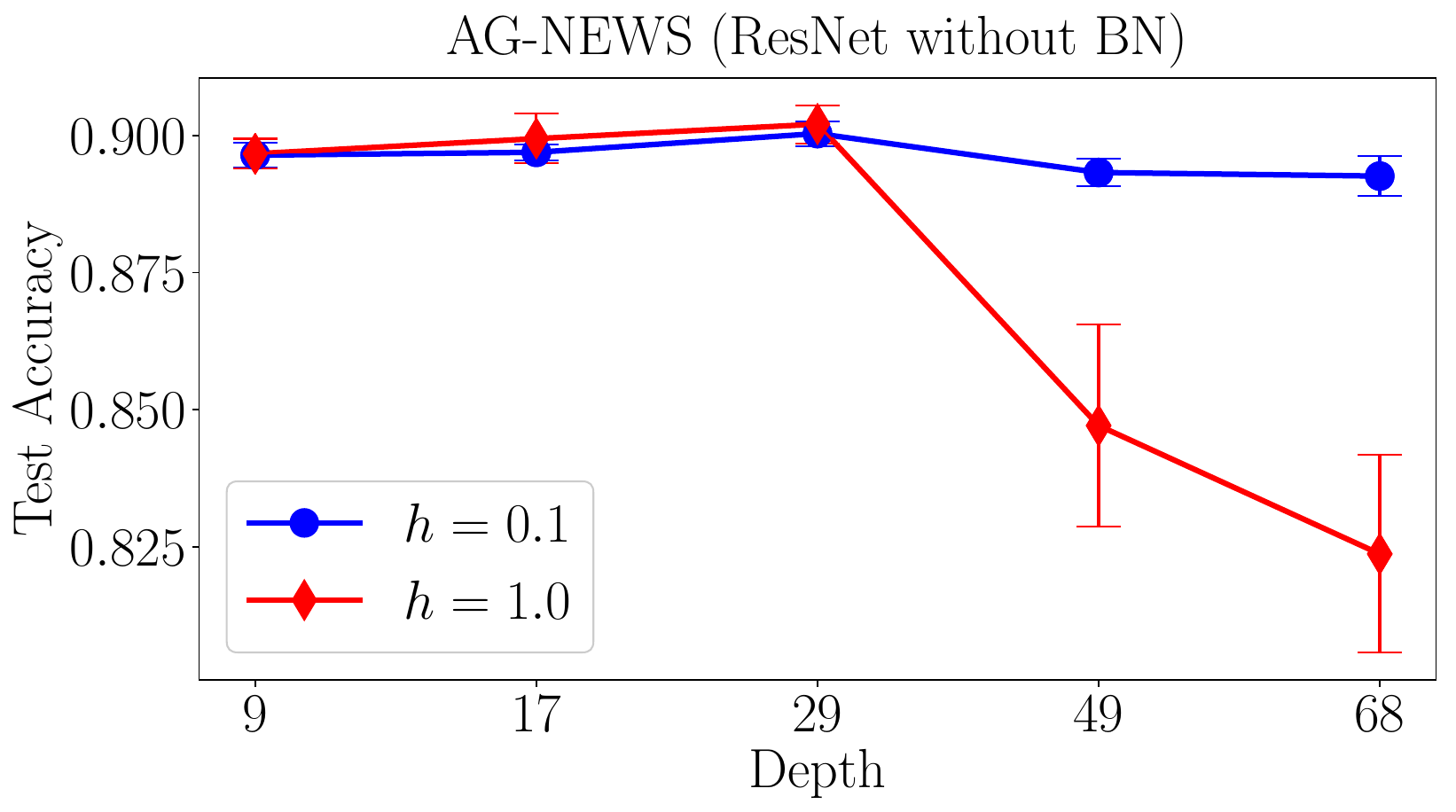}
	\caption{Training robustness comparisons of ResNet without applying BN.
	}
	\label{fig:vis_without_BN_different_depth}
\end{figure}

Fig.~\ref{fig:vis_without_BN_different_depth} shows that, without using BN, training plain ResNet ($h=1.0$) is unstable and exhibits large variance for both vision {CIFAR-10} and text {AG-NEWS} datasets, especially when the network is deep. Particularly, without using BN, even training a ResNet-110 on {CIFAR-10} fails at a high chance (2/5). However, with the reduced $h = 0.1$, training performance improves significantly and exhibits low variance. As theoretically shown in Section \ref{sec:training-robustness} reduced $h$ has beneficial effects on top of BN. This can help the back-propagated information by preventing its explosion, and thus enhance the training robustness of deep networks.

\paragraph{Small $h$ helps networks on a different optimizer - ADAM.}
\label{section:judge_small_h_helps_on_different_optimzer-adam}

\begin{figure}\centering
	\includegraphics[scale=0.25]{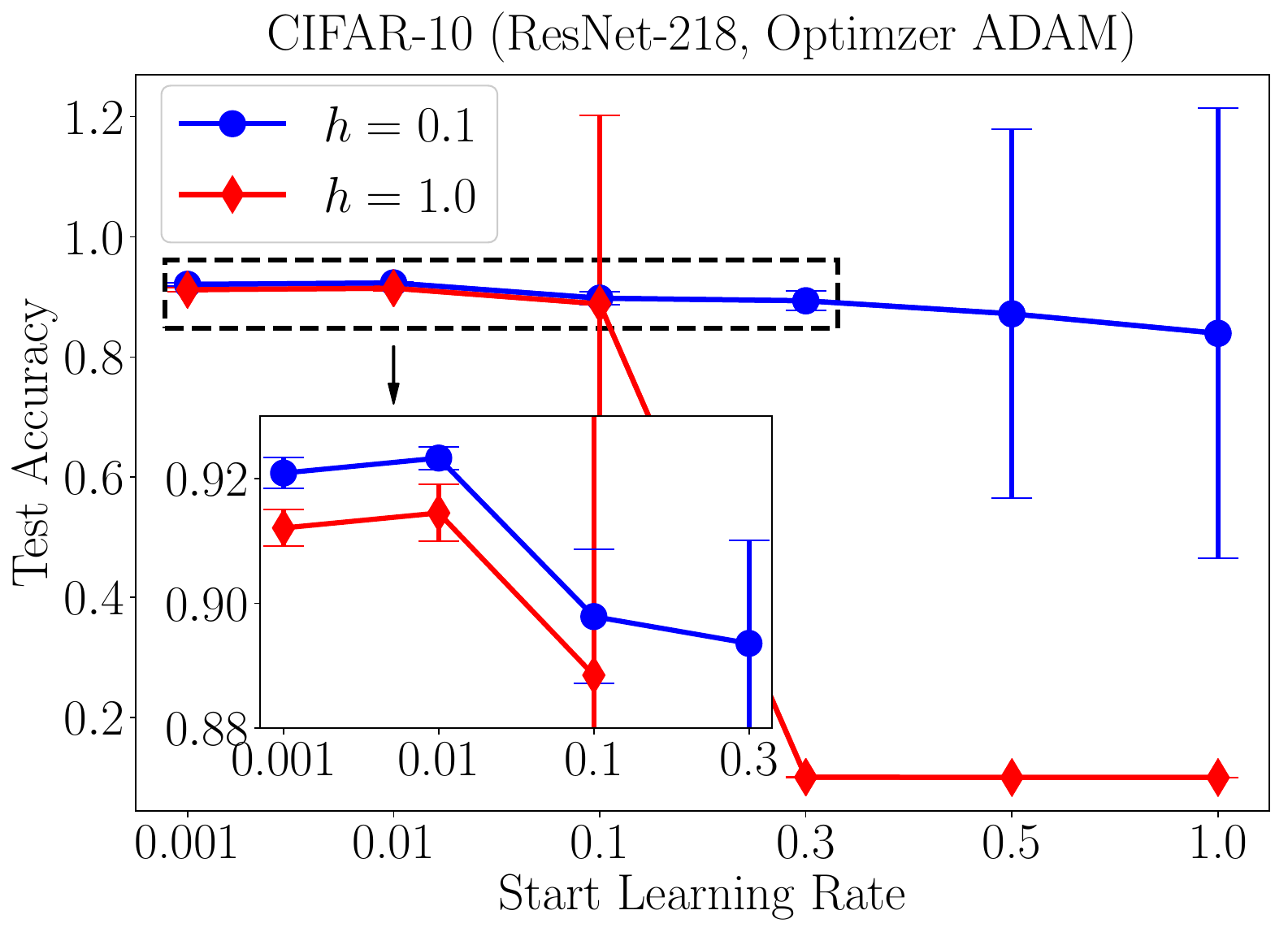}
	\caption{Training robustness comparisons of ResNet using optimizer ADAM with different learning rates.}
	\label{fig:vis_different_optimizer-adam}
\end{figure}
In order to verify that our method is robustness to different types of optimizer, in Fig.~\ref{fig:vis_different_optimizer-adam}, we train ResNet with depth 218 for CIFAR-10 dataset and compare its training performance over different start LR, but use another popular optimizer - ADAM~\cite{Kingma_Adam}. ADAM optimizer leverages the power of adaptive methods to find individual LR for each trainable parameter.

we put more comparisons graphs in Appendix~\ref{Appendix:small_h_on_adam}, i.e, ResNet with depth 44 and 110 for CIFAR-10.  
We train ResNet with 80 epochs, with different start LR divided by 10 at epochs 40 and 60 respectively.
Each configuration has 5 trails with different random seeds. Other hyperparameters remain the same. We provide median test accuracy with the standard deviation plotted as the error bar.

In Figure~\ref{fig:vis_different_optimizer-adam}, under the context of ADAM optimizer, our method (blue line) still exhibits great merits for training robustness. 
In particular, when start LR is bigger than 0.3, the existing method ($h=1.0$) fails to train ResNet, however our method ($h = 0.1$) can still stabilizing the training procedure.

Last but not least, we also compare ResNet with small $h$ and original ResNet ($h = 1.0$) over different ways of weight initialization, as detailed in Appendix~\ref{Appendix:Comparison_diff_initialization}.

\subsection{Small $h$ Improves Generalization Robustness}
\label{exp:generalization-robustness}
\begin{figure}
	\centering
	\includegraphics[scale=0.18]{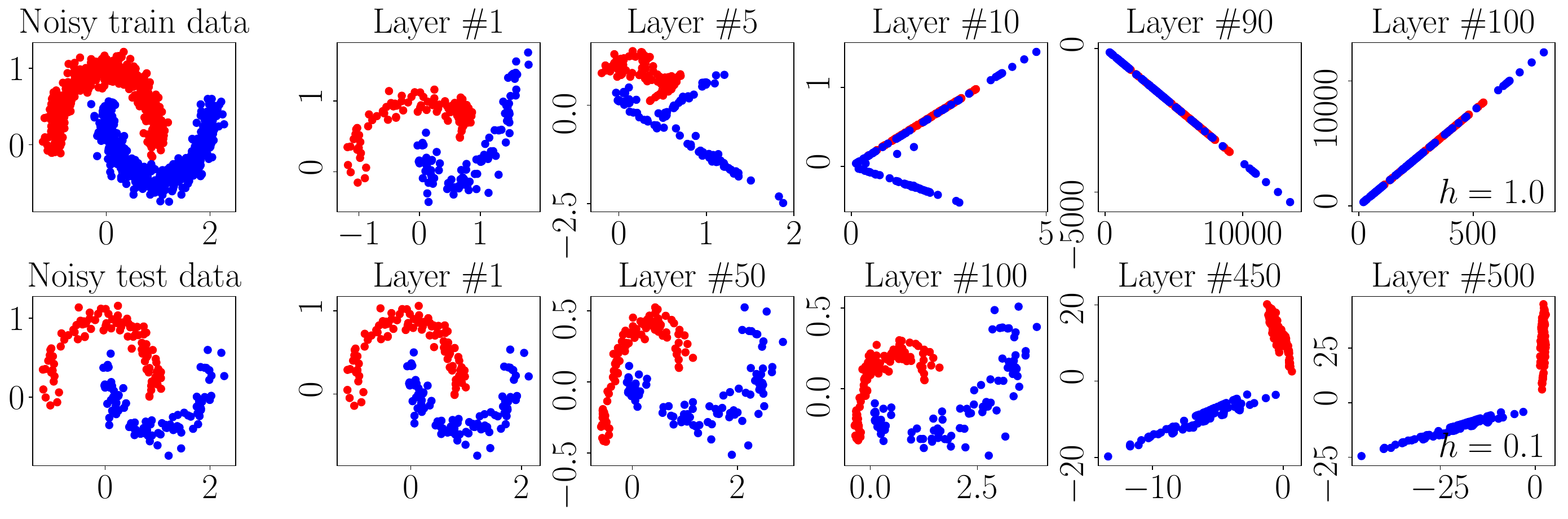}
	\caption{Illustration of feature transformations through ResNets with step factors $h = 1$ (top) and $h = 0.1$ (bottom).}
	\label{fig:moon_noisy}
\end{figure}

\textbf{Synthetic data for mitigating adverse effect of noise.} To give insights on why a small $h$ can improve the generalization robustness of ResNet, let us first consider a synthetic data example of using ResNet for a binary classification task, that is, by separating noisy ``red'' and ``blue'' points in a $2$D plane. We train a vanilla ResNet (without BN) with $h = 1$ and a ResNet with $h = 0.1$ on the training data in the top left of Fig.~\ref{fig:moon_noisy} and perform feature transformations on the test data in the bottom left of Fig.~\ref{fig:moon_noisy} using the learned ResNets. 

The series of figures at the top of Fig.~\ref{fig:moon_noisy} illustrate that the features are transformed through forward propagation in the vanilla ResNet (i.e., no BN, $h=1$), which shows that the noise in the input features leads to the mixing of red and blue points, hence sabotaging the generalization capability of ResNet. The reason for this phenomenon is that with a large $h$, the features experience violent transformations between consecutive blocks due to larger weights and the adverse effect of noise is amplified over a large depth.

On the other hand, with a small $h$, the features experience smooth transformations at every consecutive residual block and the adverse effect of the noise is thus gradually mitigated, which entails correct classifications (see the series of figures at the bottom of Fig.~\ref{fig:moon_noisy}). As mentioned in Section~\ref{sec:generalization-robustness}, a small $h$ can help to mitigate the adverse effect of noise in the input features. 
With a small $h$, the noise amplification is effectively limited over a large depth. \\
\textbf{Real-world data for mitigating adverse effect of noise.} 
To verify the effectiveness of a small $h$ on the generalization robustness of ResNet,
we train on noisy input data with varying noise levels; illustrations of noisy input are detailed in Appendix~\ref{Appendix:vis_noisy_input}. We provide the test accuracy of ResNets with depths $218$ and $49$ on the clean input data of CIFAR-10 and AG-NEWS datasets, respectively. For the AG-NEWS dataset, LR is fixed to $0.01$. We compare test accuracy of the ResNet with reduced step factor $h= 0.1$ and the original ResNet (i.e., $h= 1.0$)\textcolor{red}{;} other hyperparameters remain the same. Each configuration has $5$ trials with different random seeds. The standard deviation is plotted as an error bar.
\begin{figure}
	\centering
	\includegraphics[scale=0.14]{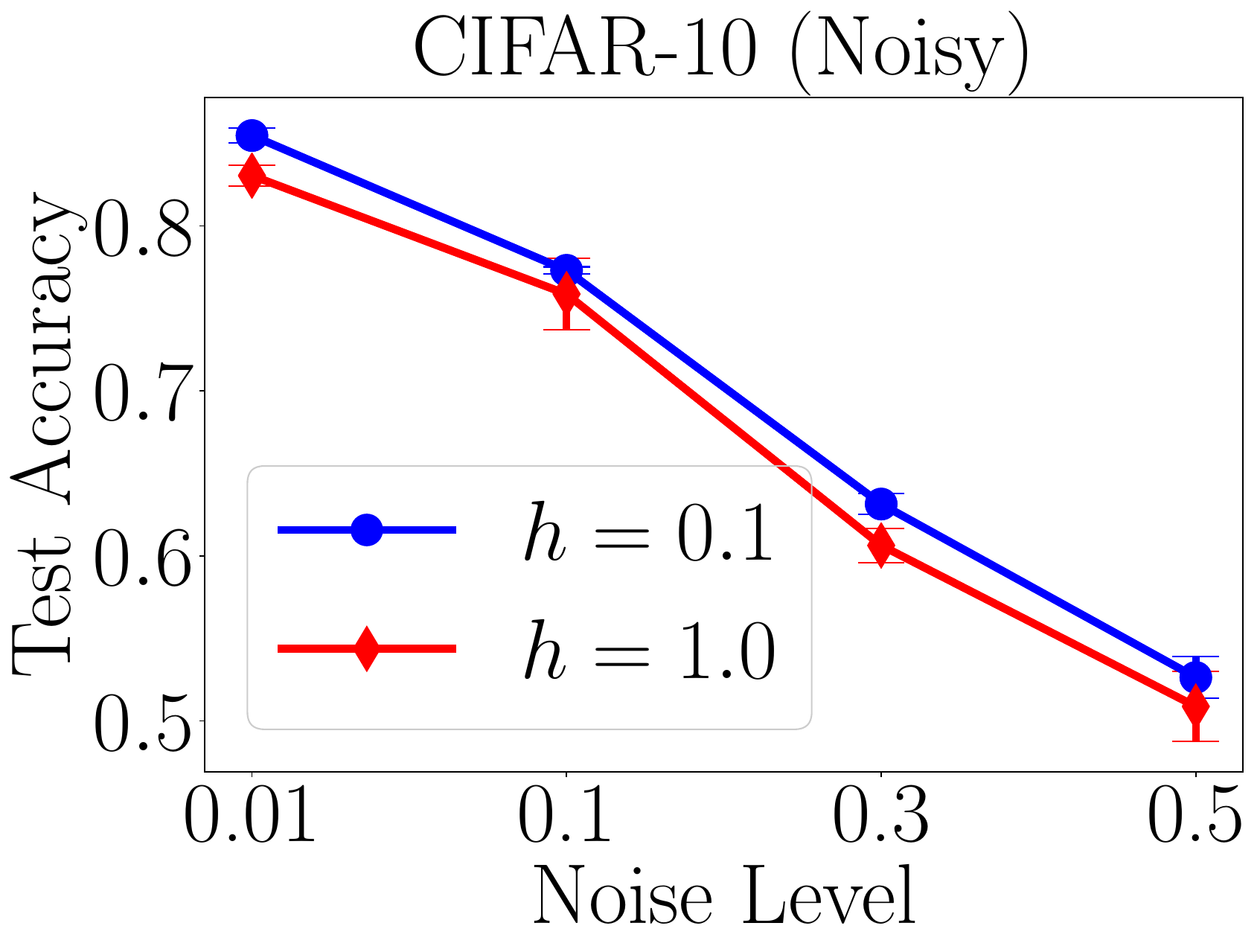}
	\includegraphics[scale=0.14]{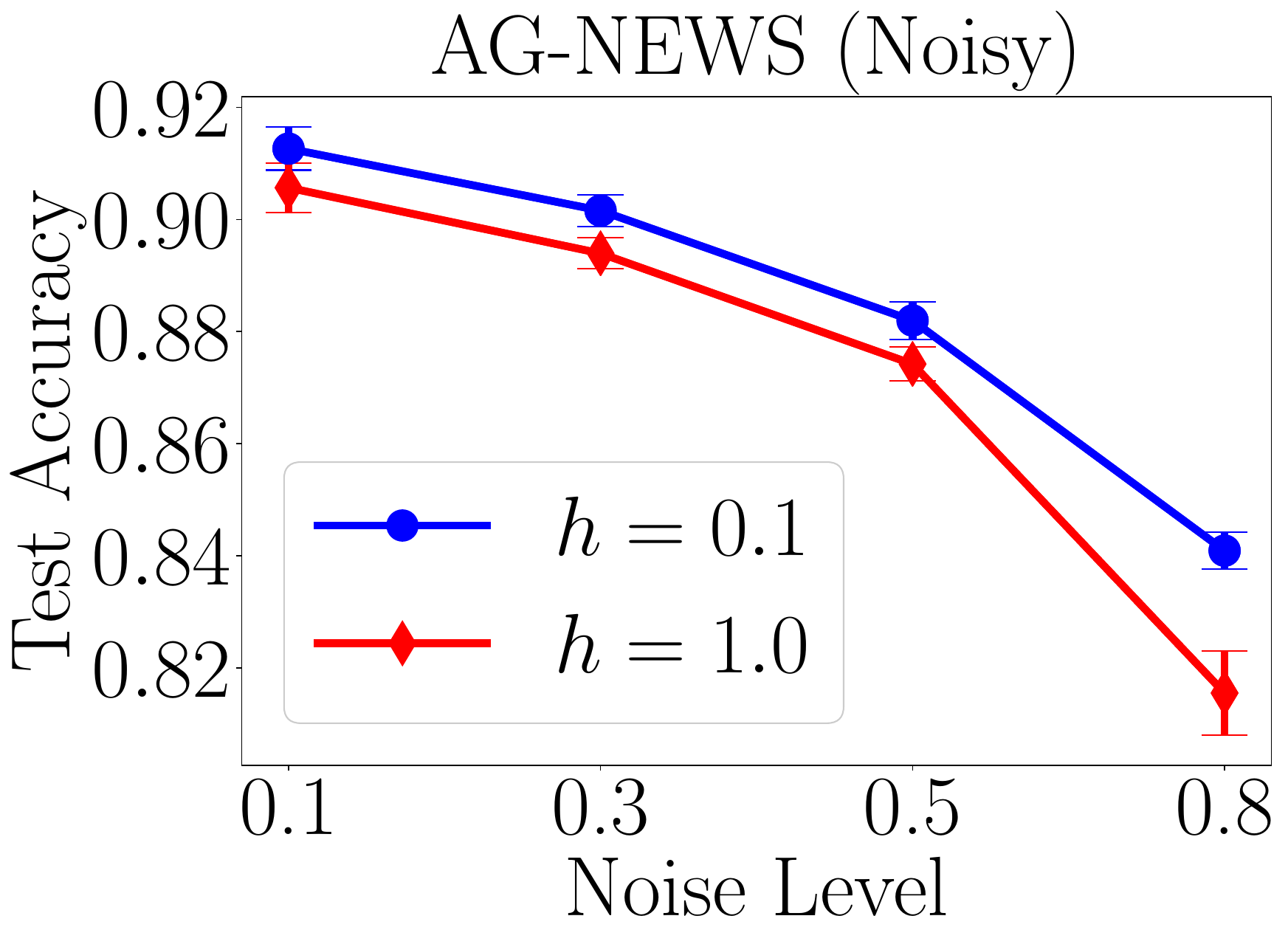}
	\caption{Effect of $h = 1$ and $h = 0.1$ on generalization robustness of ResNet.}
	\label{fig:real-world-noisy-input}
\end{figure}

Fig.~\ref{fig:real-world-noisy-input} shows that at different noise levels, our ResNet with reduced $h = 0.1$ consistently outperforms the original ResNet (i.e., $h = 1$). 
It can also be observed that our ResNet with reduced $h=0.1$ has a smaller variance than its counterpart under different noise levels. 
Hence, our ResNet with $h=0.1$ is robust to training on noisy input by mitigating the adverse effect of noise. By making smooth transformations, it gradually mitigates the adverse effect of noise in the input features along the forward propagation of ResNet. So, our ResNet with $h=0.1$ offers better generalization robustness.
\subsection{How to Select Step Factor $h$}\label{exp:selection-of-h}
%
We perform a grid search of $h$ from $0.001$ to $20$ for the CIFAR-10 dataset and from $0.001$ to $1.0$ for the AG-NEWS dataset to optimize $h$.
We train ResNets over different step factors $h$.
We provide the median test accuracy with each configuration having $5$ trials with different random seeds. The standard deviation is plotted as an error bar.
\begin{figure}
\begin{tabular}{c}
	\includegraphics[scale=0.24]{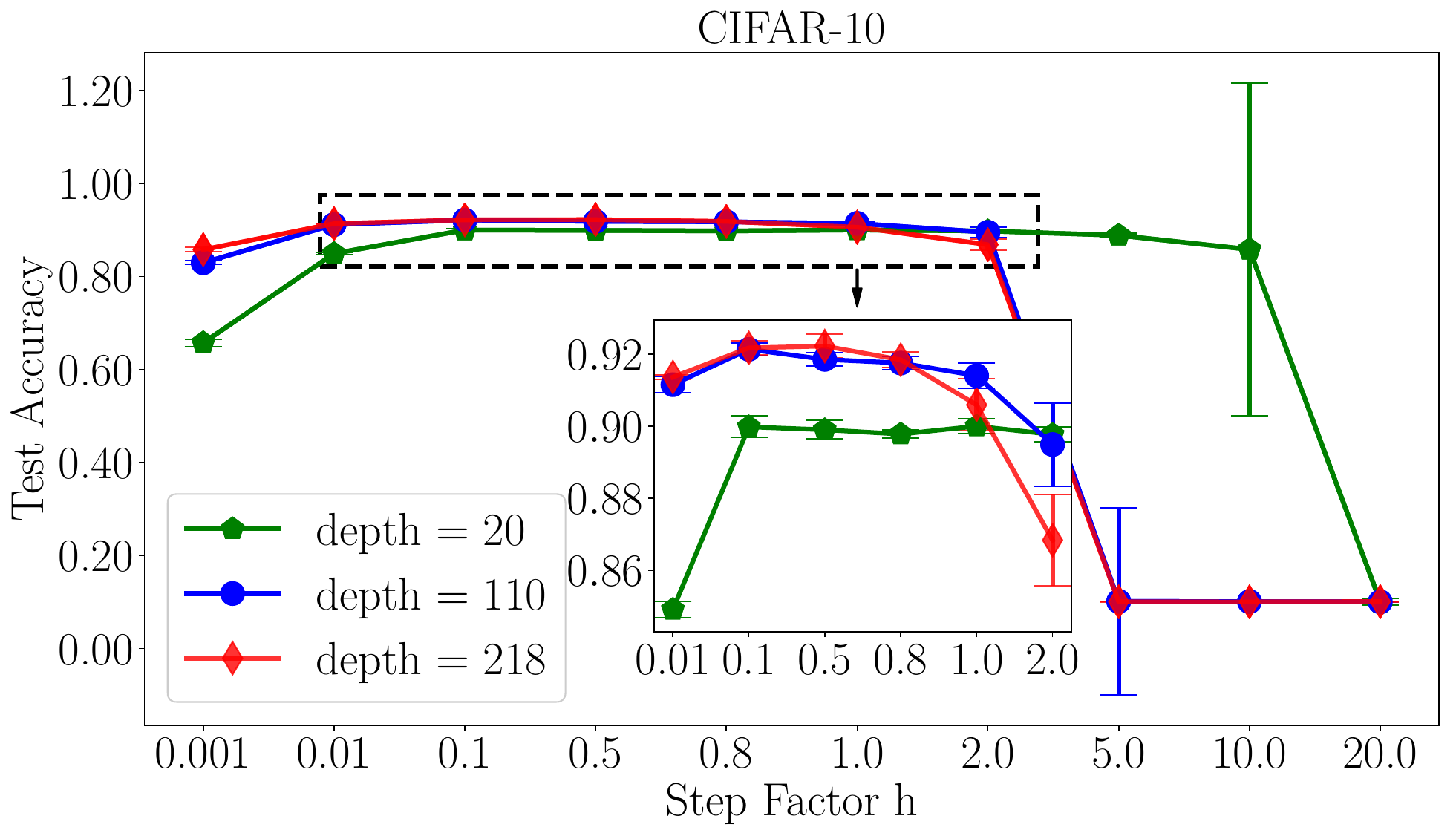}\\
	\includegraphics[scale=0.24]{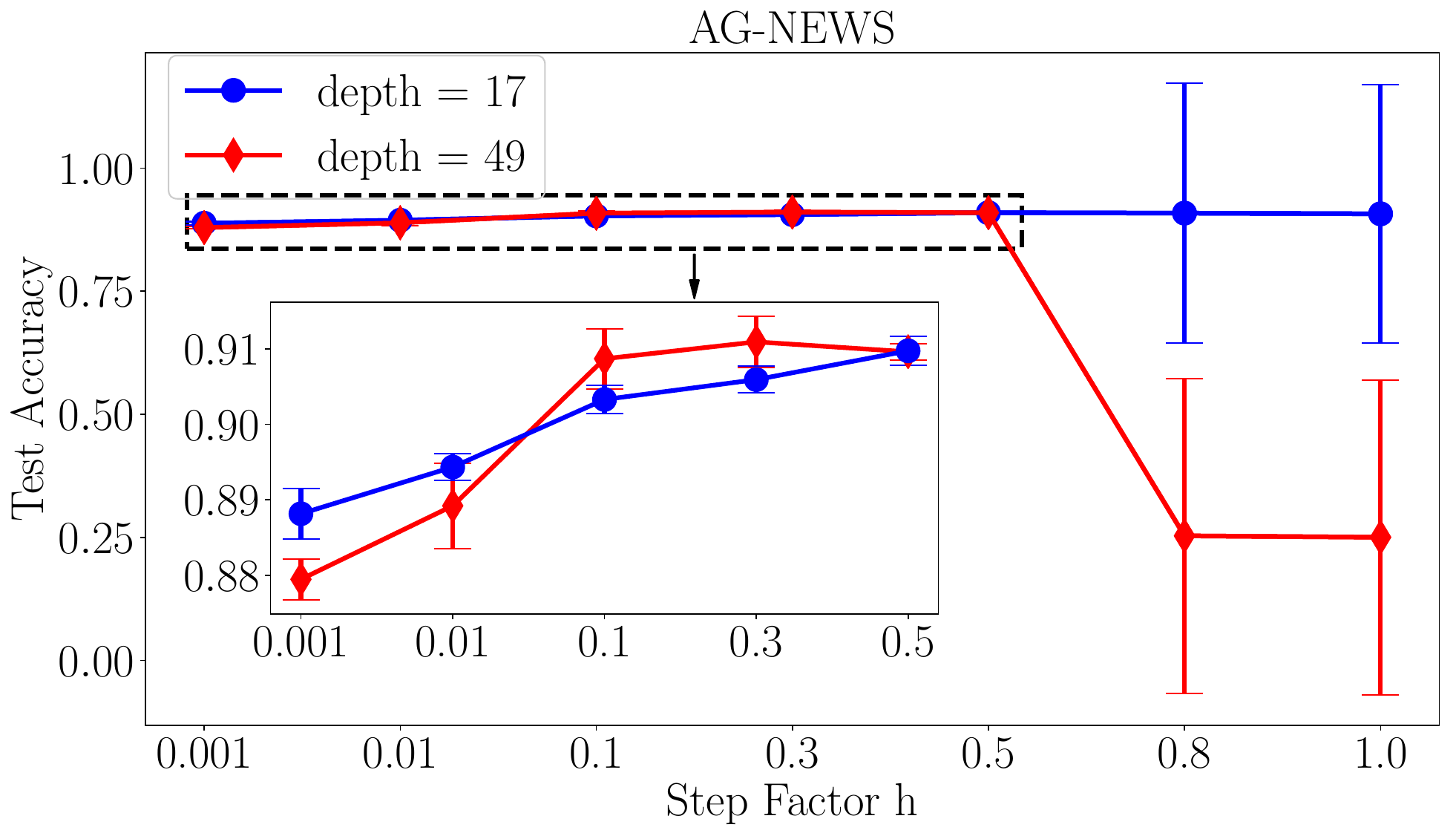}
\end{tabular}	
	\caption{Selection of step factor $h$ for ResNet with varying depths.}
	\label{fig:training-robust-h}
\end{figure}

Fig.~\ref{fig:training-robust-h} shows that the optimal $h$ is near $0.1$ for ResNet-$110$ and ResNet-$218$ and in \textcolor{red}{$[0.1, 2]$} for ResNet-$20$ for the CIFAR-10 dataset. For the AG-NEWS dataset, the optimal $h$ is near $0.3$ for ResNet-$49$ and $0.5$ for ResNet-$17$. 

When $h$ is small, the error bar is small as well (i.e., performance variance is small). 
In comparison, when $h$ is very large (e.g., $h\geq 5$ for ResNet-$110$ for the CIFAR-10 dataset), 
training becomes unstable and often fails. 
In ResNet-$17$ and ResNet-$49$ for the AG-NEWS dataset, when $h \geq 0.8$, $3$ out of $5$ training procedures fail in our experiments.

In addition, we  observe that when $h$ is very small (e.g., $h = 0.001$), generalization performance degrades even though the training variance is very small. This confirms our claim in  Section~\ref{sec:generalization-robustness} that an overly small $h$ would smooth out the feature transformations. 

To summarize, the guideline for the selection of $h$ is that it should be small but not too small.
Our experiments reveal that for a deep ResNet, $h$ should be smaller (e.g., $h = 0.1$ for ResNet-$218$ for CIFAR-10 dataset) while for a shallow ResNet, a larger $h$ (e.g., $h = 2.0$ for ResNet-$20$ for CIFAR-10 dataset) can be tolerated. In addition, once the depth of ResNet is fixed, the chosen $h$ should not be too small (e.g., $h=0.001$ and $0.01$) so as not to smooth out the feature transformations. 
\section{Conclusion}
This paper describes a simple yet principled approach to boost the robustness of ResNet. Motivated by the dynamical system perspective, we characterize a ResNet using an explicit Euler method. This consequently allows us to exploit the step factor $h$ in the Euler method to control the robustness of ResNet in both its training and generalization. 
We prove that a small step factor $h$ can benefit its training and generalization robustness during backpropagation and forward propagation, respectively.
Empirical evaluation on real-world datasets corroborates our analytical findings that a small $h$ can indeed improve both its training and generalization robustness. For future work, we plan to explore several promising directions: (a) how to transfer the experience of a small $h$ to other network structures (e.g., RNN for natural language processing~\cite{cho2014learning}),
(b) how to handle the noisy target output labels~\cite{han2018co}, and
(c) other means to choose the step size $h$ (e.g., using Bayesian optimization \cite{zhongxiang,daxberger2017distributed,hoang2018decentralized,ling16,zhang2017information,yehong}).

\clearpage

\bibliographystyle{named}
\bibliography{ijcai19.bib}

\newpage
\appendix

\onecolumn
\section{Proof of Proposition~\ref{yesyesyes}} ~\label{Appendix: Proof_of_Prop_1}
From~\eqref{back_prop},
\begin{equation*}
	\displaystyle\frac{\partial L}{\partial \mathbf{x}_0} =  \displaystyle \frac{\partial L}{\partial \mathbf{x}_D} \left(\mathbf{1}+ h \frac{\partial}{\partial \mathbf{x}_0} \sum_{i=0}^{D-1} \mathcal{F} (\mathbf{x}_i)  \right).
\end{equation*}	
It follows that 
\begin{equation*}
\hspace{-1.9mm}
\begin{array}{l}	
	\displaystyle \left\lVert \frac{\partial L}{\partial \mathbf{x}_0} \right\rVert\vspace{1mm}\\ 
	\leq  \displaystyle \left\lVert \frac{\partial L}{\partial \mathbf{x}_D} \right\rVert \left(1+ h \left\lVert\frac{\partial}{\partial \mathbf{x}_0} \sum_{i=0}^{D-1} \mathcal{F} (\mathbf{x}_i) \right\rVert \right)  \vspace{1mm}\\ 
	=\displaystyle \left\lVert \frac{\partial L}{\partial \mathbf{x}_D} \right\rVert   \left(1 + h \left \lVert    
	 \frac{\partial \mathcal{F}(\mathbf{x}_0) } {\partial {\mathbf{x}_0}}   +  \frac{\partial \mathcal{F}(\mathbf{x}_1)}{\partial {\mathbf{x}_0}} + \ldots +  \frac{\partial \mathcal{F}(\mathbf{x}_{D-1})}{\partial {\mathbf{x}_0}}
	\right \rVert \right)  \vspace{1mm}
	 \\
	 = \displaystyle \left\lVert \frac{\partial L}{\partial \mathbf{x}_D} \right\rVert   \left(1 + h \left \lVert    
	 \frac{\partial \mathcal{F}(\mathbf{x}_0) }{\partial {\mathbf{x}_0}}   +  \frac{\partial \mathcal{F}(\mathbf{x}_1)}{\partial {\mathbf{x}_1}} \frac{\partial \mathbf{x}_1}{\partial {\mathbf{x}_0}} + \ldots +  \frac{\partial \mathcal{F}(\mathbf{x}_{D-1})}{\partial {\mathbf{x}_{D-1}}  } 
	 \frac{\partial \mathbf{x}_{D-1} }  {\partial \mathbf{x}_{D-2}  }   
	 \ldots
	  \frac{\partial \mathbf{x}_{1} }  {\partial \mathbf{x}_{0}  }   
	\right  \rVert \right)  \vspace{1mm}
	 \\
	  \leq  \displaystyle \left\lVert \frac{\partial L}{\partial \mathbf{x}_D} \right\rVert   \left(1 + h \left \lVert    
	 \frac{\partial \mathcal{F}(\mathbf{x}_0) }{\partial {\mathbf{x}_0}} \right \rVert  + h\left \lVert \frac{\partial \mathcal{F}(\mathbf{x}_1)}{\partial {\mathbf{x}_1}} \frac{\partial \mathbf{x}_1}{\partial {\mathbf{x}_0}}\right \rVert + \ldots +  	
	h\left  \lVert \frac{\partial \mathcal{F}(\mathbf{x}_{D-1})}{\partial {\mathbf{x}_{D-1}}  } 
 \frac{\partial \mathbf{x}_{D-1} }  {\partial \mathbf{x}_{D-2}  }   
	 \ldots
	 \frac{\partial \mathbf{x}_{1} }  {\partial \mathbf{x}_{0}  }   
	\right \rVert \right)  \vspace{1mm}
	 \\
	  = \displaystyle \left\lVert \frac{\partial L}{\partial \mathbf{x}_D} \right\rVert   \left(1 + h\left \lVert    
	 \frac{\partial \mathcal{F}(\mathbf{x}_0)}{\partial {\mathbf{x}_0}} \right \rVert   + 
	 h\left \lVert \frac{\partial \mathcal{F}(\mathbf{x}_1)}{\partial {\mathbf{x}_1}} \left(\mathbf{1}+ h \frac{\partial \mathcal{F}(\mathbf{x}_0) }{\partial {\mathbf{x}_0}} \right) \right\rVert + \ldots + 
	h \left \lVert \frac{\partial \mathcal{F}(\mathbf{x}_{D-1})}{\partial {\mathbf{x}_{D-1}}  } 
	 \left(\mathbf{1}+ h \frac{\partial \mathcal{F}(\mathbf{x}_{D-2}) }{\partial {\mathbf{x}_{D-2}}} \right)
	 \ldots
	\left(\mathbf{1}+ h \frac{\partial \mathcal{F}(\mathbf{x}_0) }{\partial {\mathbf{x}_0}} \right)
	\right  \rVert  \right)  \vspace{1mm}
	 \\
	 \leq\displaystyle \left\lVert \frac{\partial L}{\partial \mathbf{x}_D}  \right \rVert  \left(1+ h  \mathcal{W}+ h \mathcal{W}(\sqrt{d}+h\mathcal{W}) + \ldots +h \mathcal{W}(\sqrt{d}+h\mathcal{W})^{D-1} \right)\vspace{1mm} \\ 
	 = \displaystyle  \left \lVert \frac{\partial L}{\partial \mathbf{x}_D} \right \rVert \left (1+ h \mathcal{W}\sum_{i=0}^{D-1}    (\sqrt{d}+h\mathcal{W})^i   \right )  \vspace{1mm} \\ 
	   = \displaystyle  \left \lVert \frac{\partial L}{\partial \mathbf{x}_D} \right \rVert \left (1+ h \mathcal{W}  \frac{(\sqrt{d}+h\mathcal{W} )^ {D}-1}{\sqrt{d}+h\mathcal{W}-1}
	    \right )  \vspace{1mm} \\ 
	    \leq \displaystyle  \left \lVert \frac{\partial L}{\partial \mathbf{x}_D} \right \rVert \left (1+ h \mathcal{W}  \frac{(\sqrt{d}+h\mathcal{W} )^ {D}-1}{h\mathcal{W}}
	    \right )  \vspace{1mm} \\
	 = \displaystyle \left \lVert \frac{\partial L}{\partial \mathbf{x}_D} \right \rVert    (\sqrt{d}+h\mathcal{W})^D \ .
\end{array}
\end{equation*}
The first inequality is due to triangle inequality and submultiplicativity of the matrix norm.
The second equality follows from the chain rule. 
The second inequality is due to triangle inequality.
The third equality follows from substituting $N=n+1$ into~\eqref{recursive_resblock} and differentiating~\eqref{recursive_resblock} with respect to $\mathbf{x}_n$.
The third inequality follows from submultiplicativity of the matrix norm, triangle inequality, and the assumption of $\lVert{\partial \mathcal{F}}(\mathbf{x}_i)/{\partial \mathbf{x}_i} \rVert\leq\mathcal{W}$ for $i=0,\ldots,D-1$.
The second last equality follows from geometric progression. 

\clearpage 

\section{Small h Sabilizes Gradients and Encourages Smaller Weights.}
\label{Appendix:judge_small_h_stabilizing_gradients_weights}
\begin{figure} [h] ~\centering
	\includegraphics[scale=0.16]{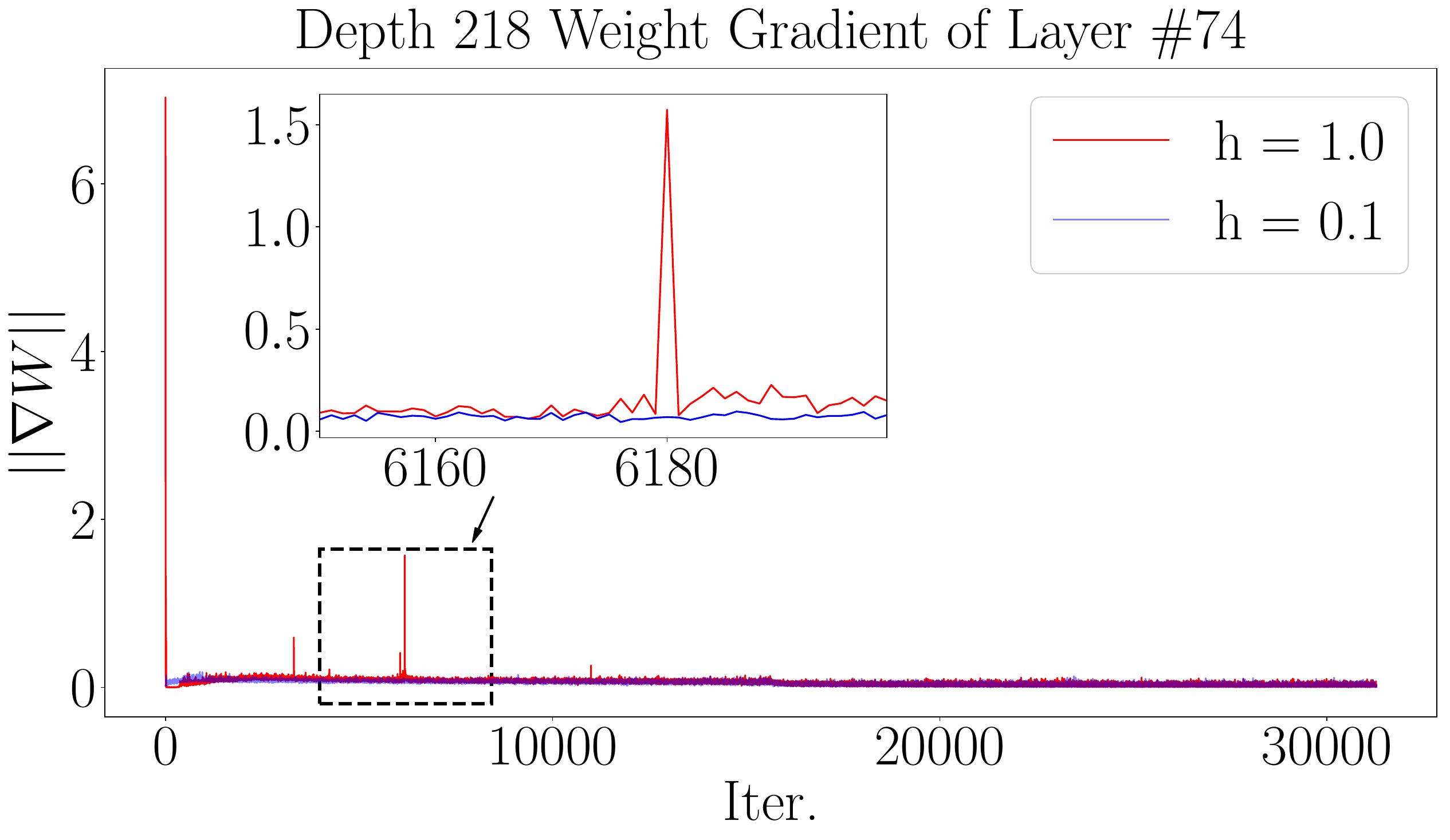}
	\includegraphics[scale=0.16]{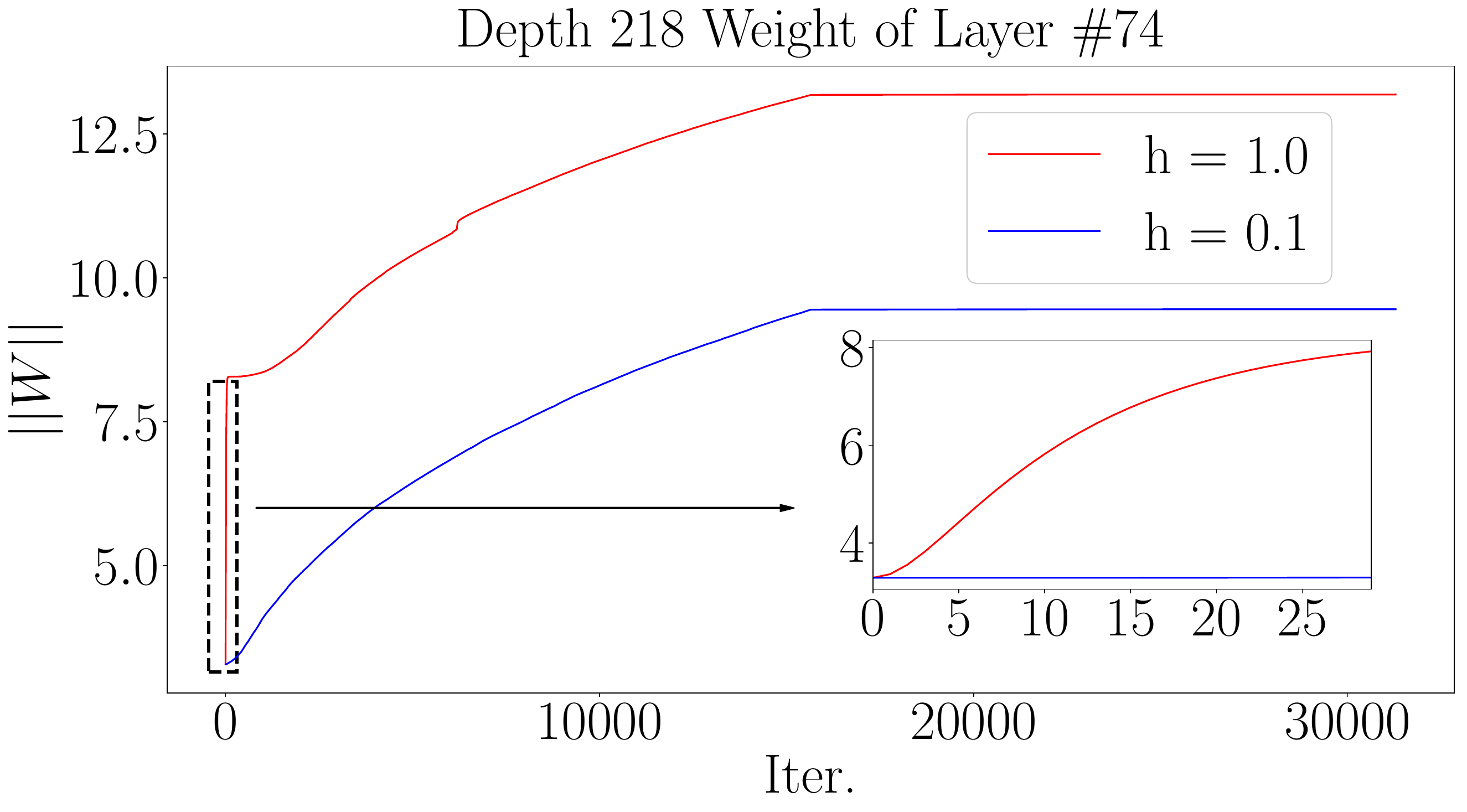}\\
	\includegraphics[scale=0.16]{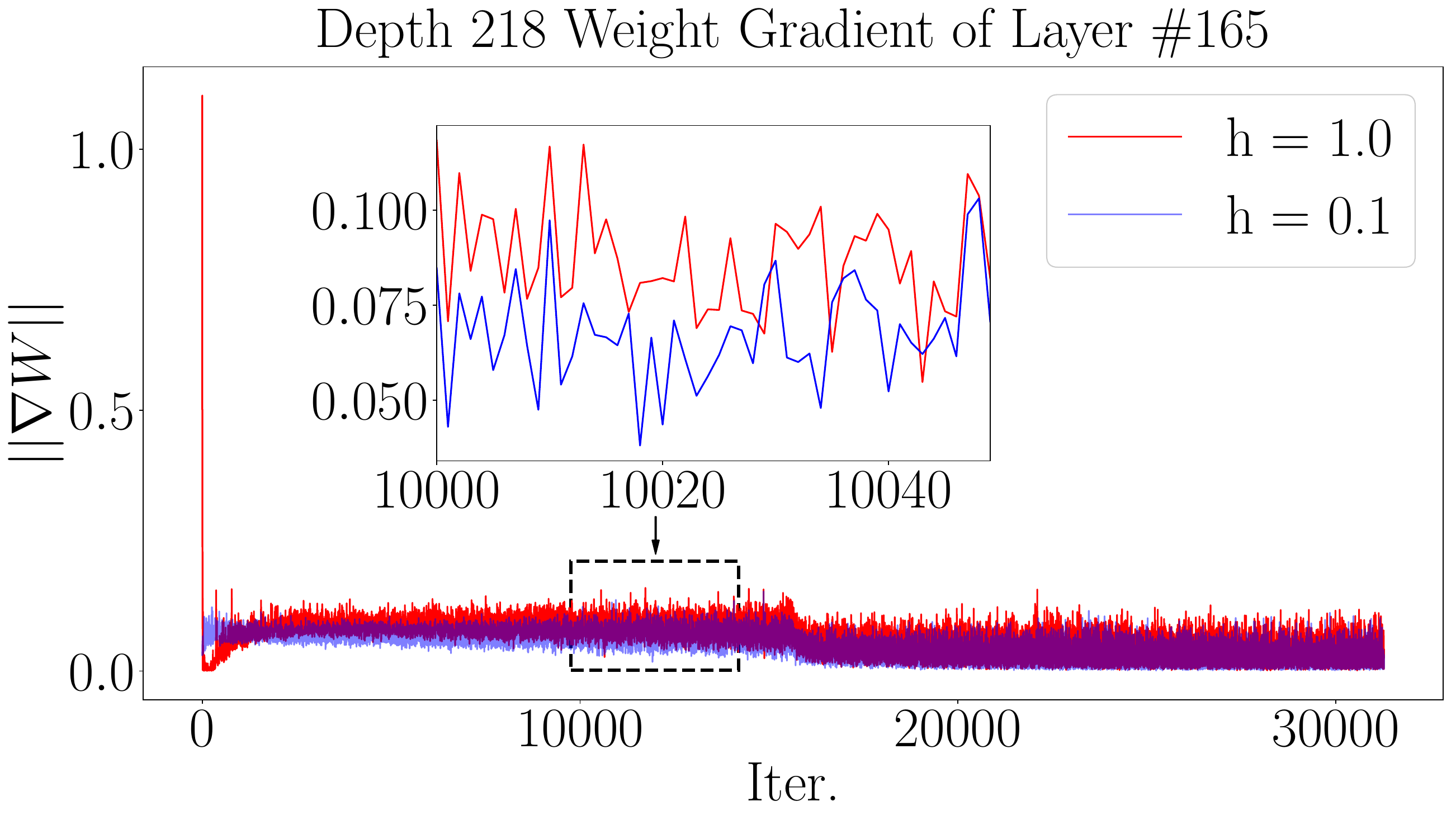}
	\includegraphics[scale=0.16]{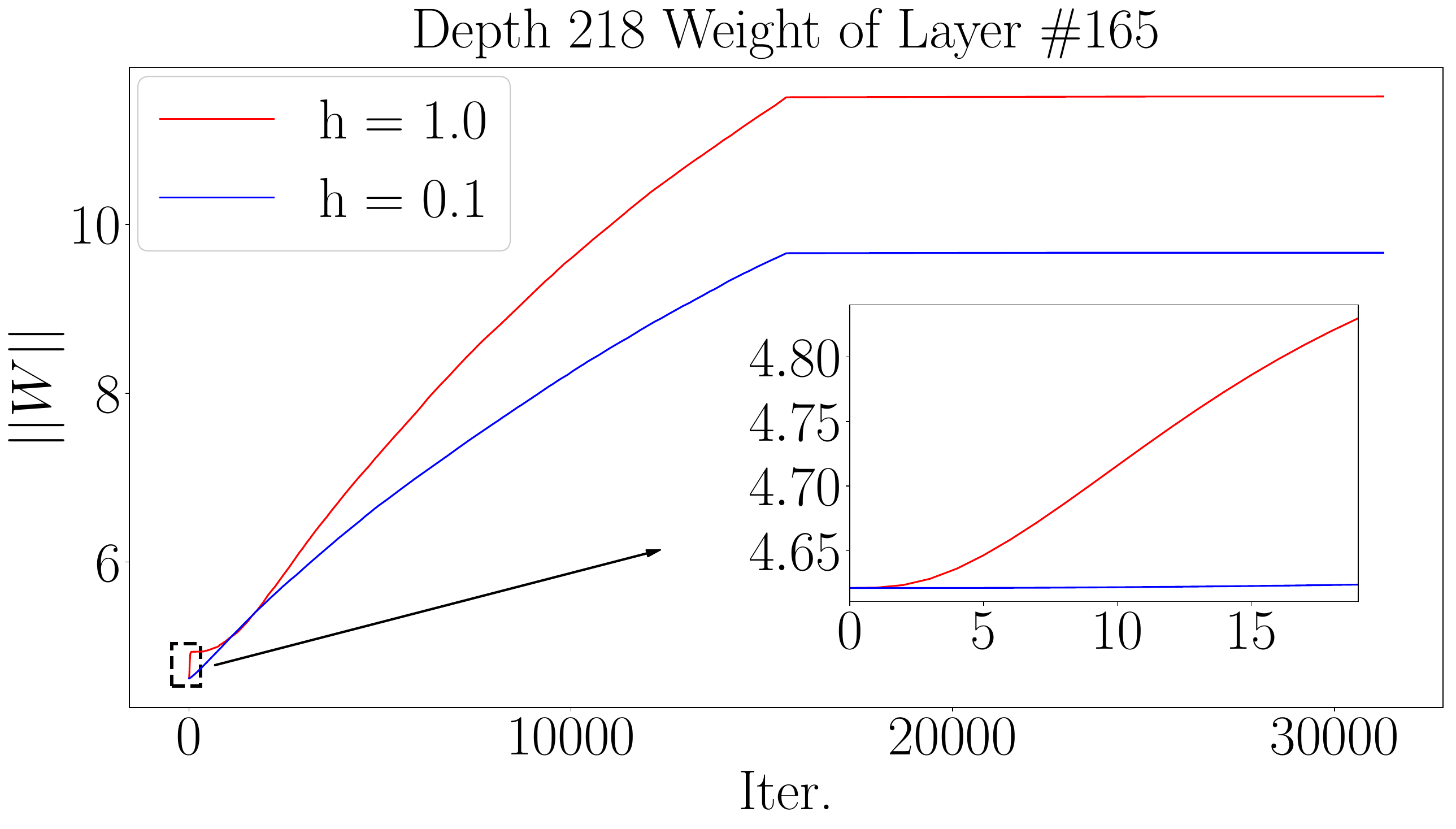}\\
	\includegraphics[scale=0.16]{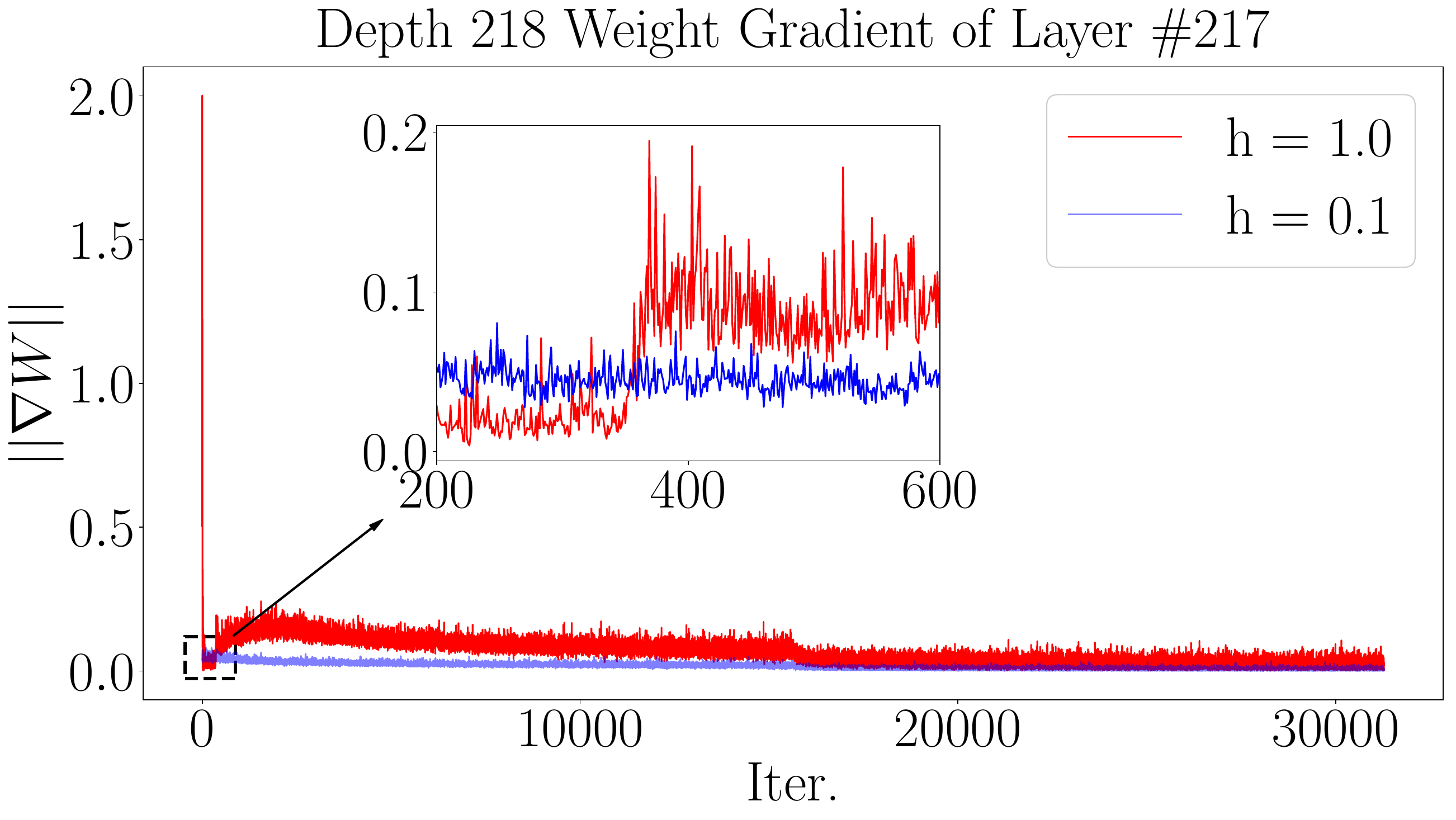}
	\includegraphics[scale=0.16]{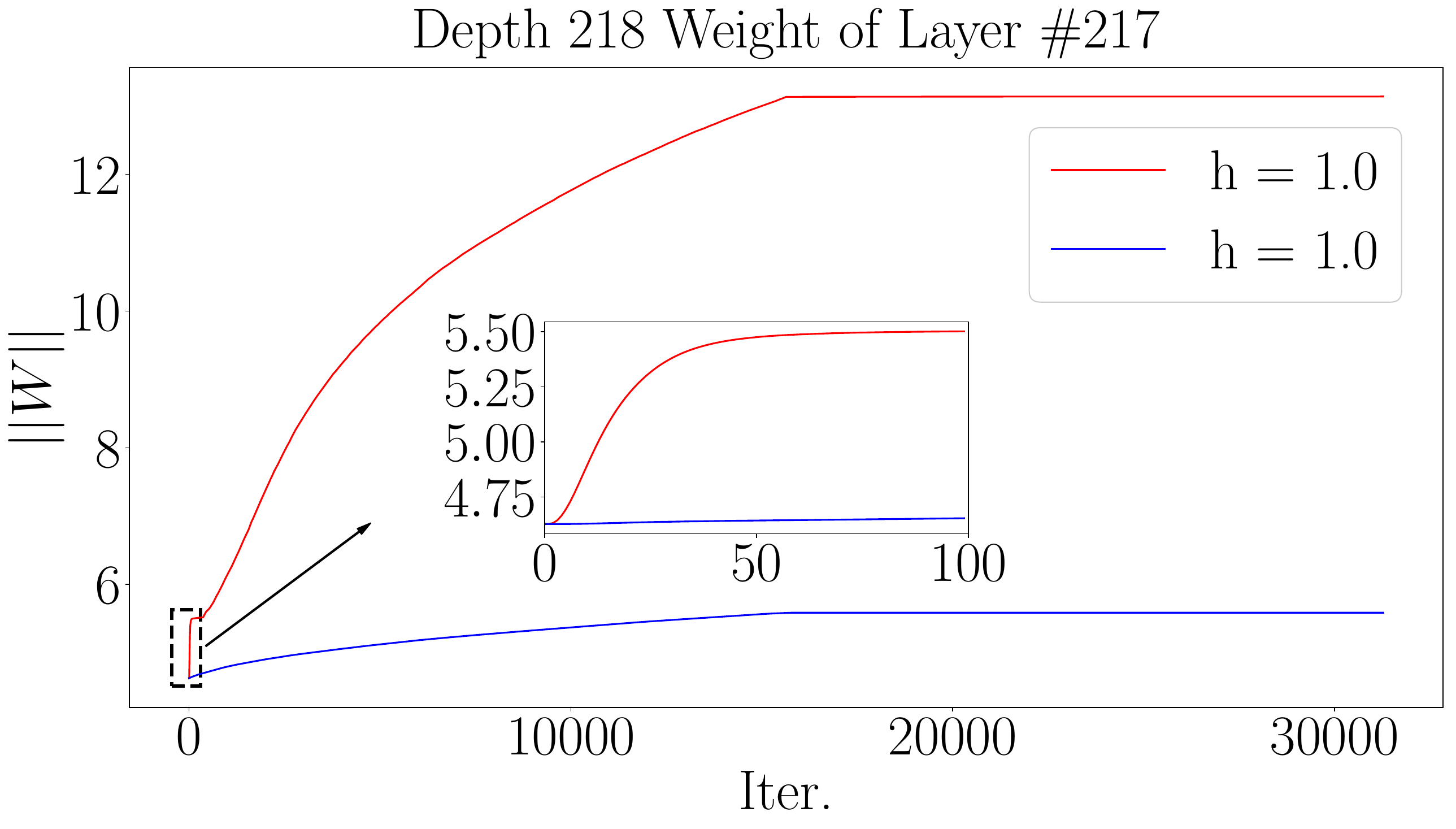}
	\caption{Gradients and weights dynamics over training iterations of ResNet-218 on {CIFAR-10}.
	}
	\label{judge_gradients_and_weights-Res218}
\end{figure}
\clearpage
\begin{figure} [h] ~\centering
	\includegraphics[scale=0.16]{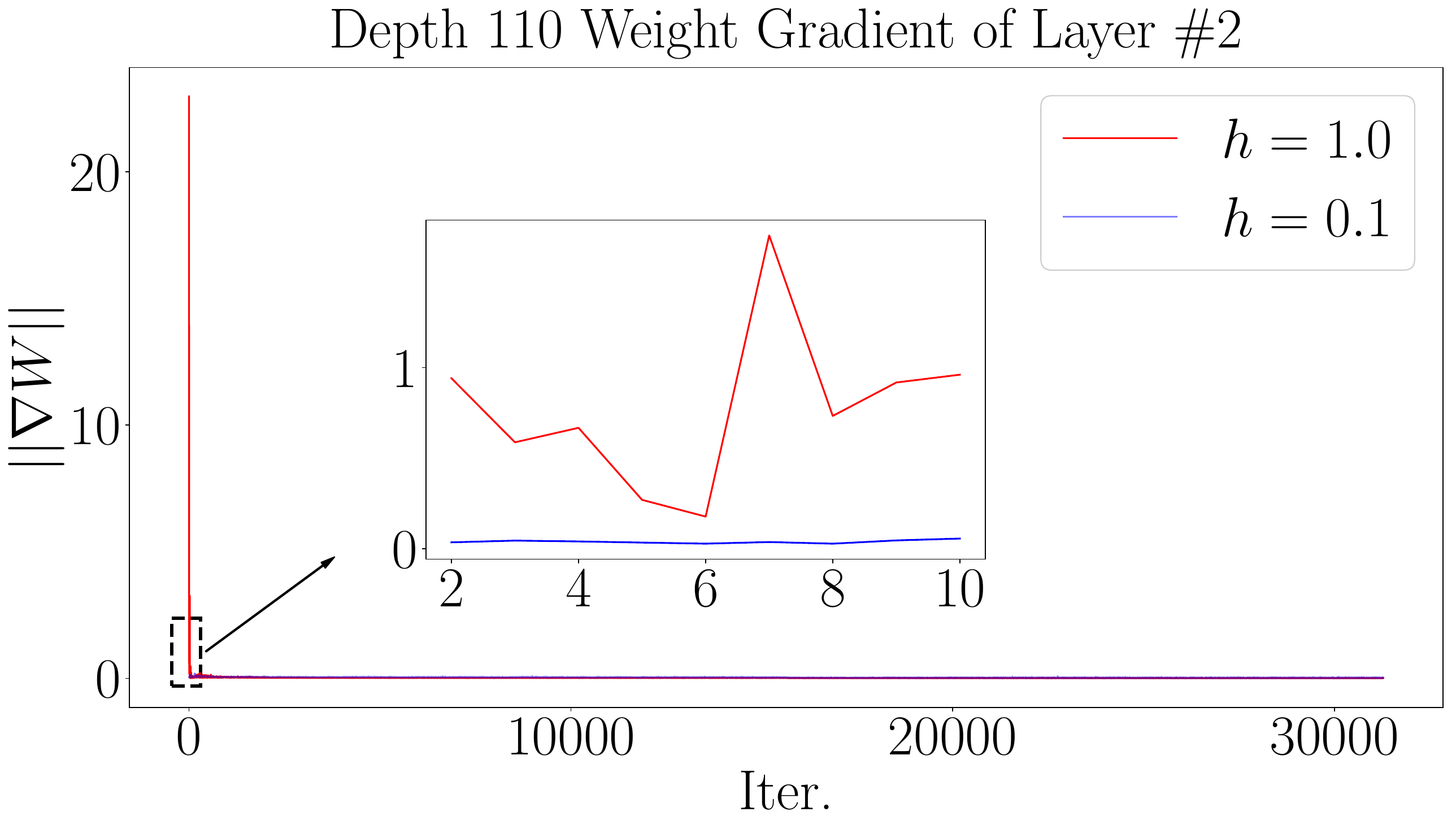}
	\includegraphics[scale=0.16]{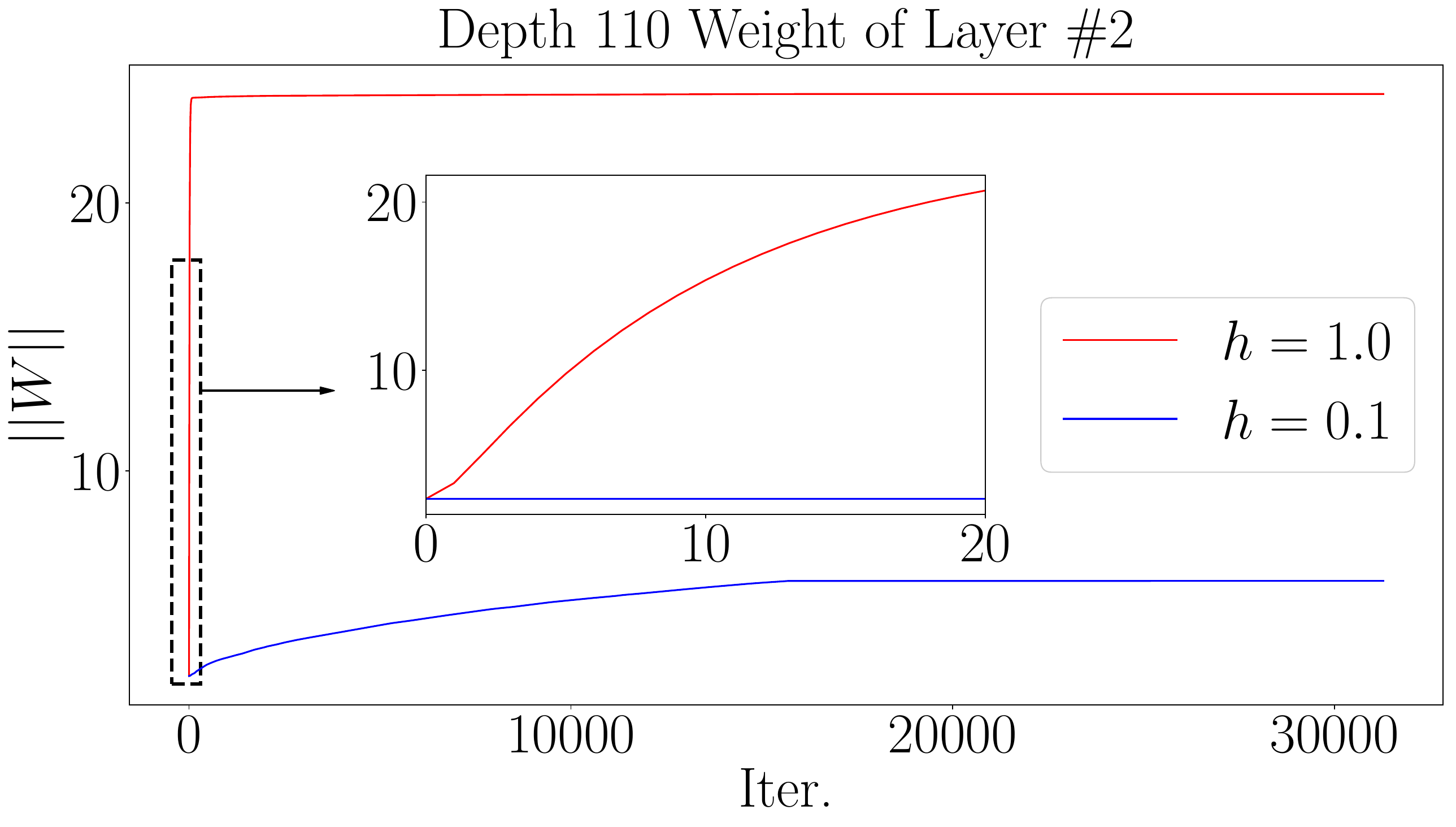}\\
	\includegraphics[scale=0.16]{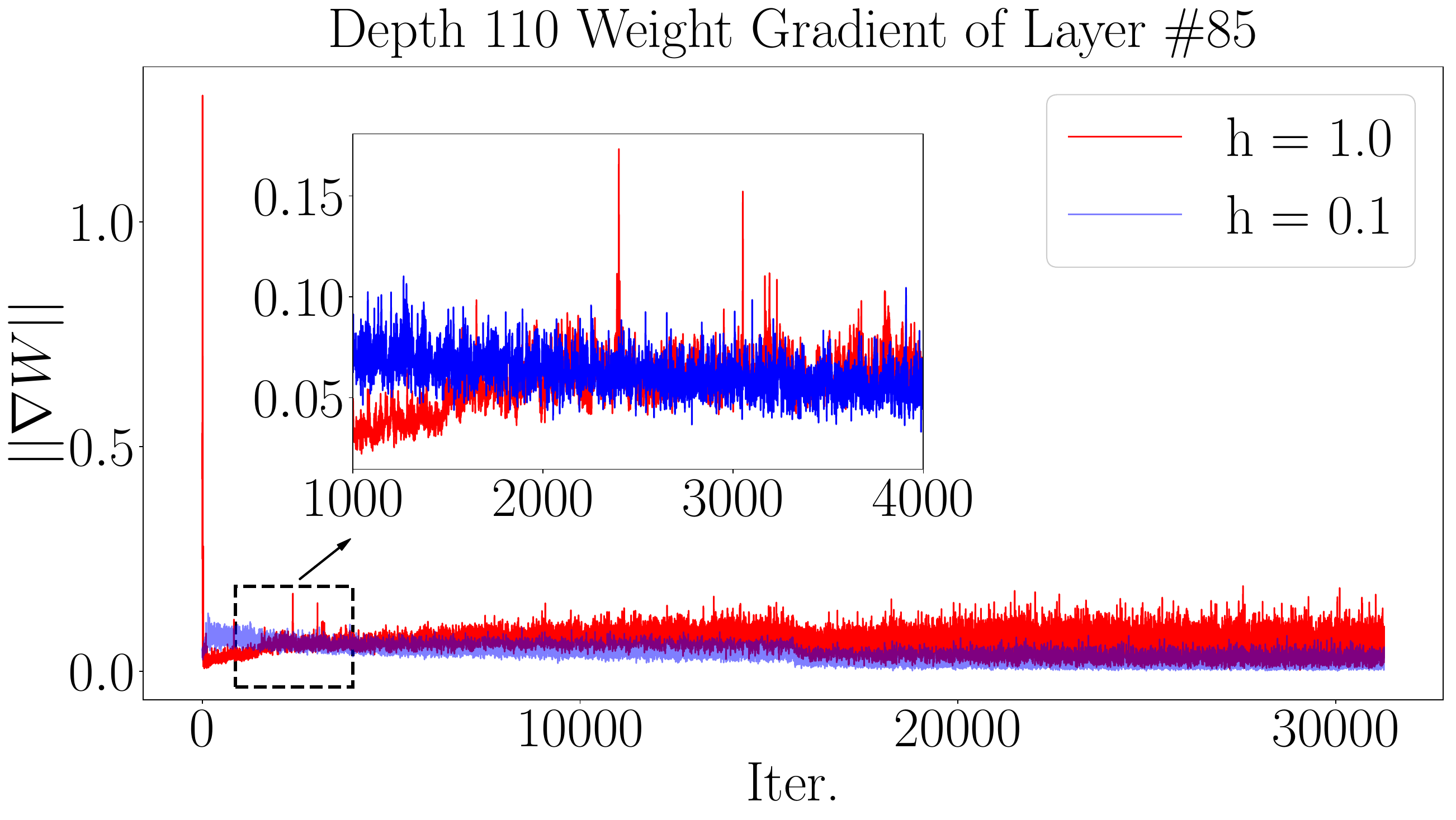}
	\includegraphics[scale=0.16]{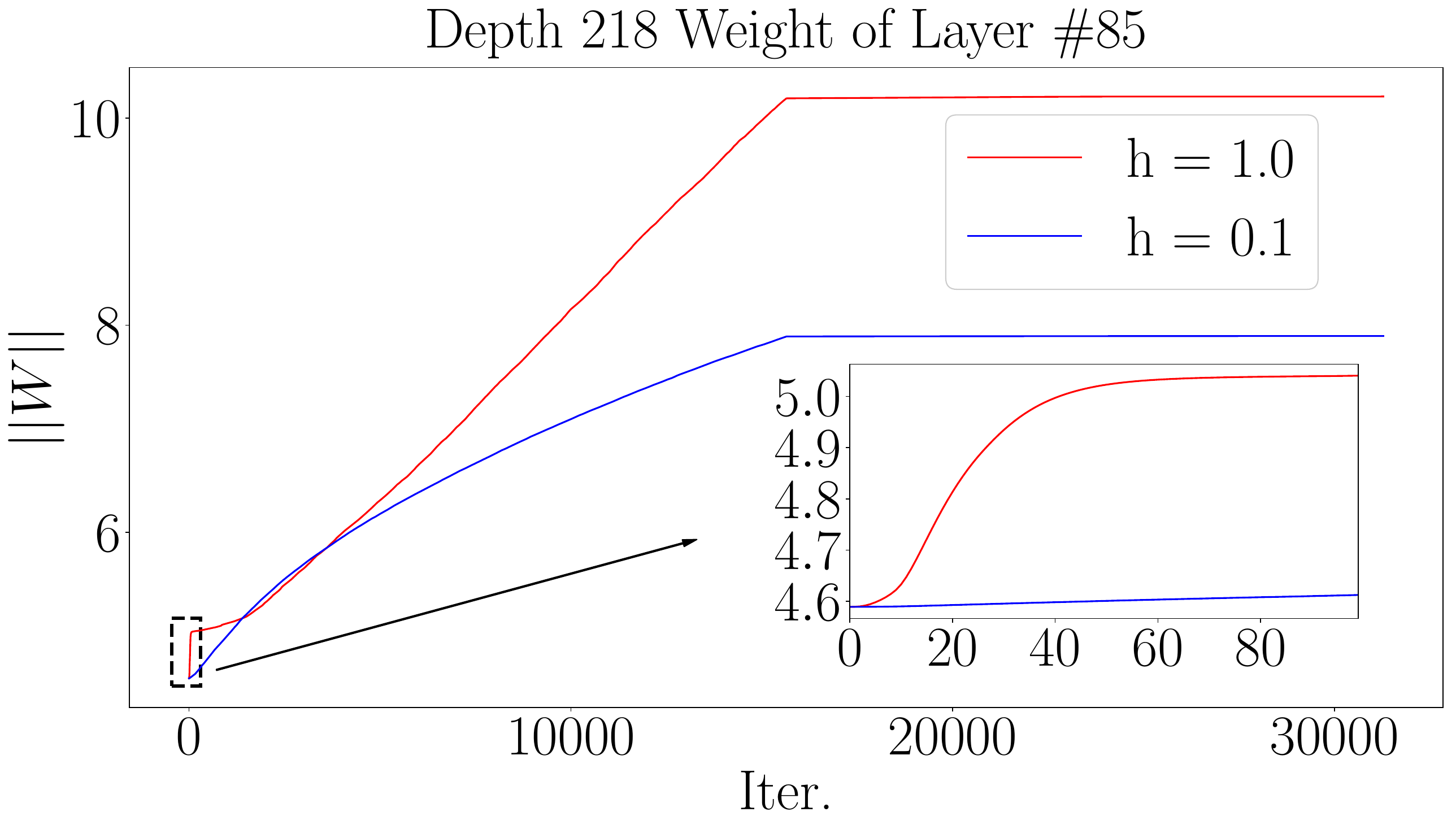}\\
	\includegraphics[scale=0.16]{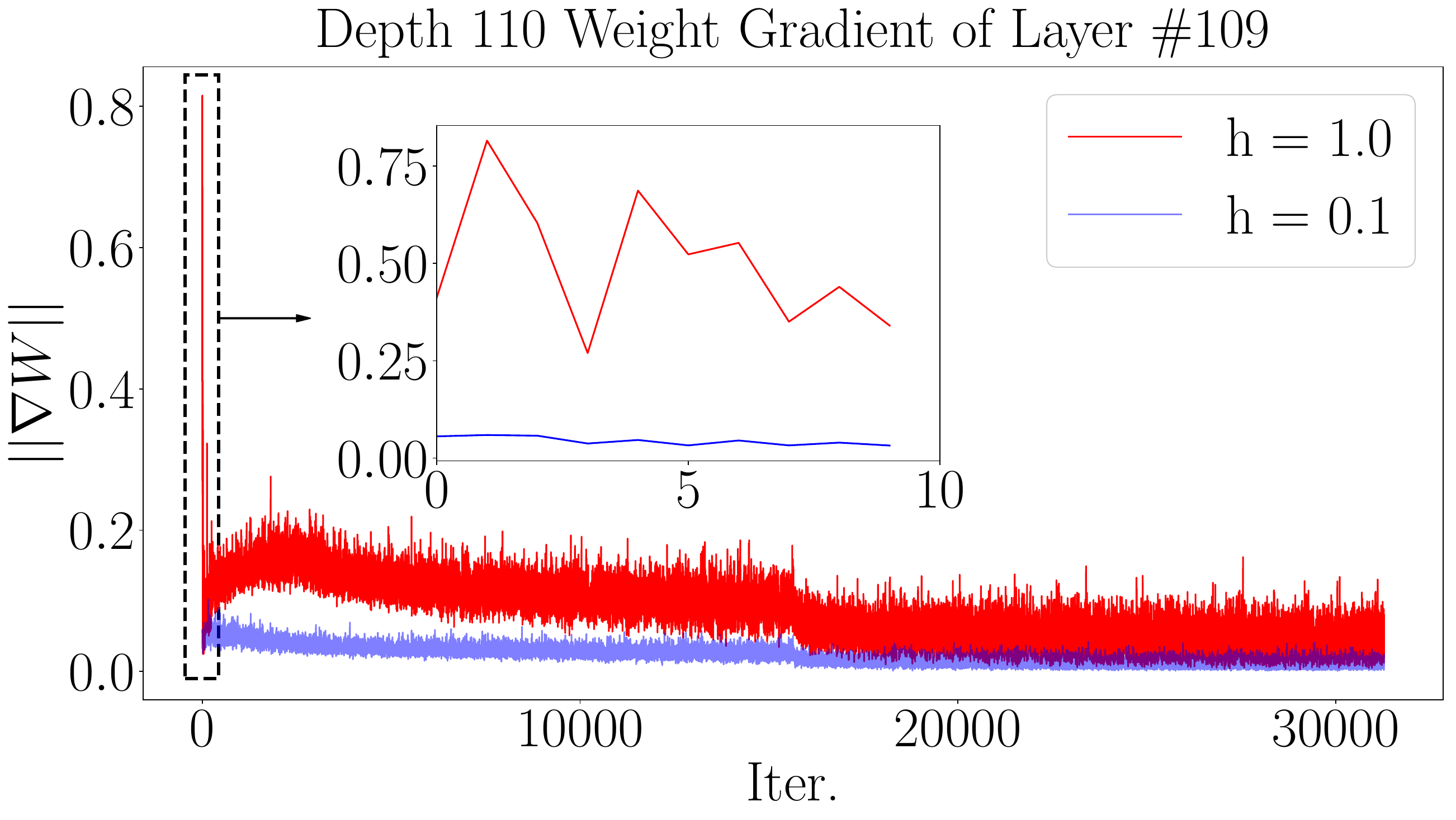}
	\includegraphics[scale=0.16]{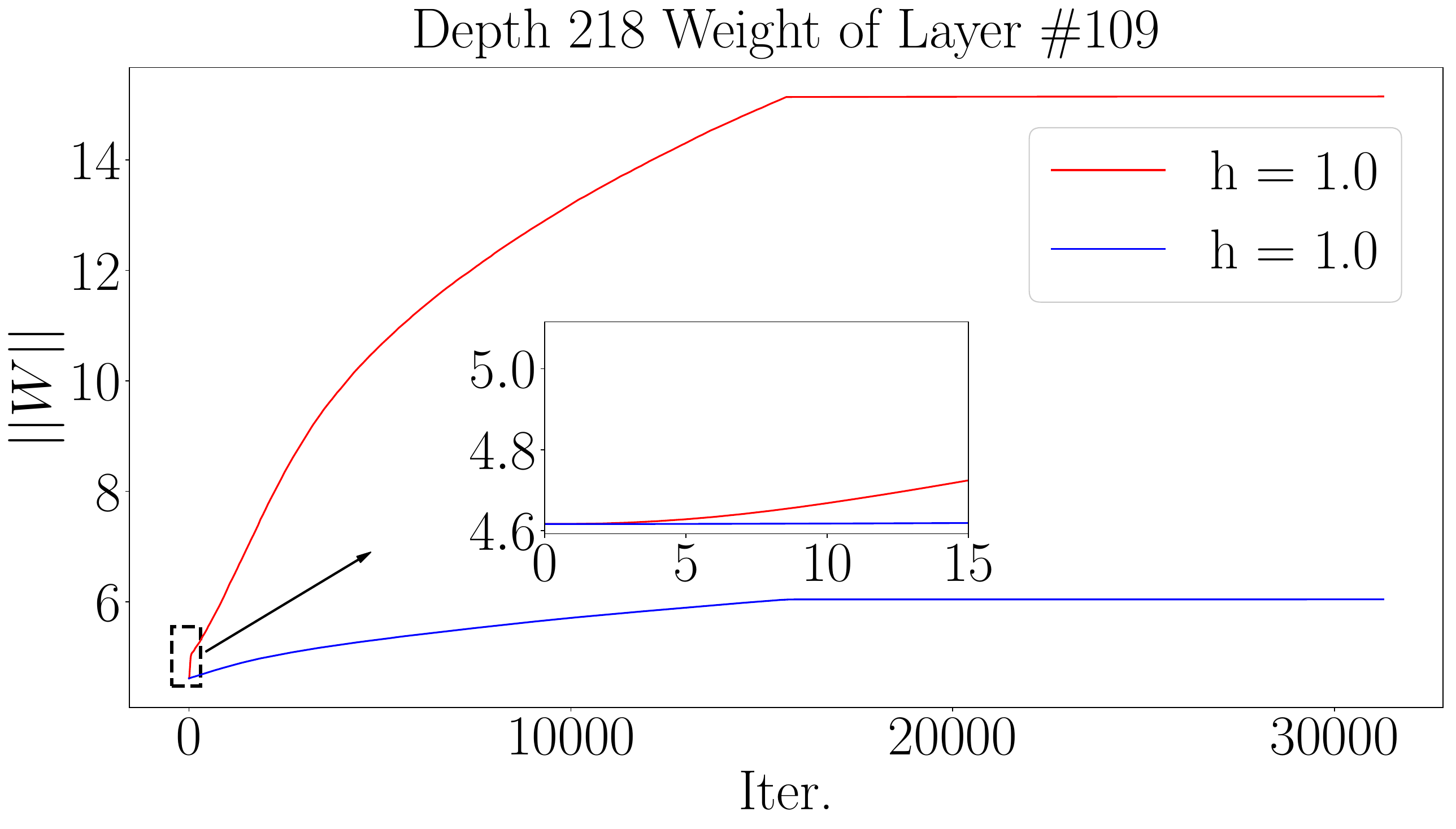}
	\caption{Gradients and weights dynamics over training iterations of ResNet-110 on {CIFAR-10}.
	}
	\label{judge_gradients_and_weights-Res218}
\end{figure}
\clearpage

\section{Small $h$ Enables Larger Learning Rate (LR)}
\label{Appendix:judge_small_h_enable_larger_LR}
\begin{figure}[h!]\centering
	\includegraphics[scale=0.25]{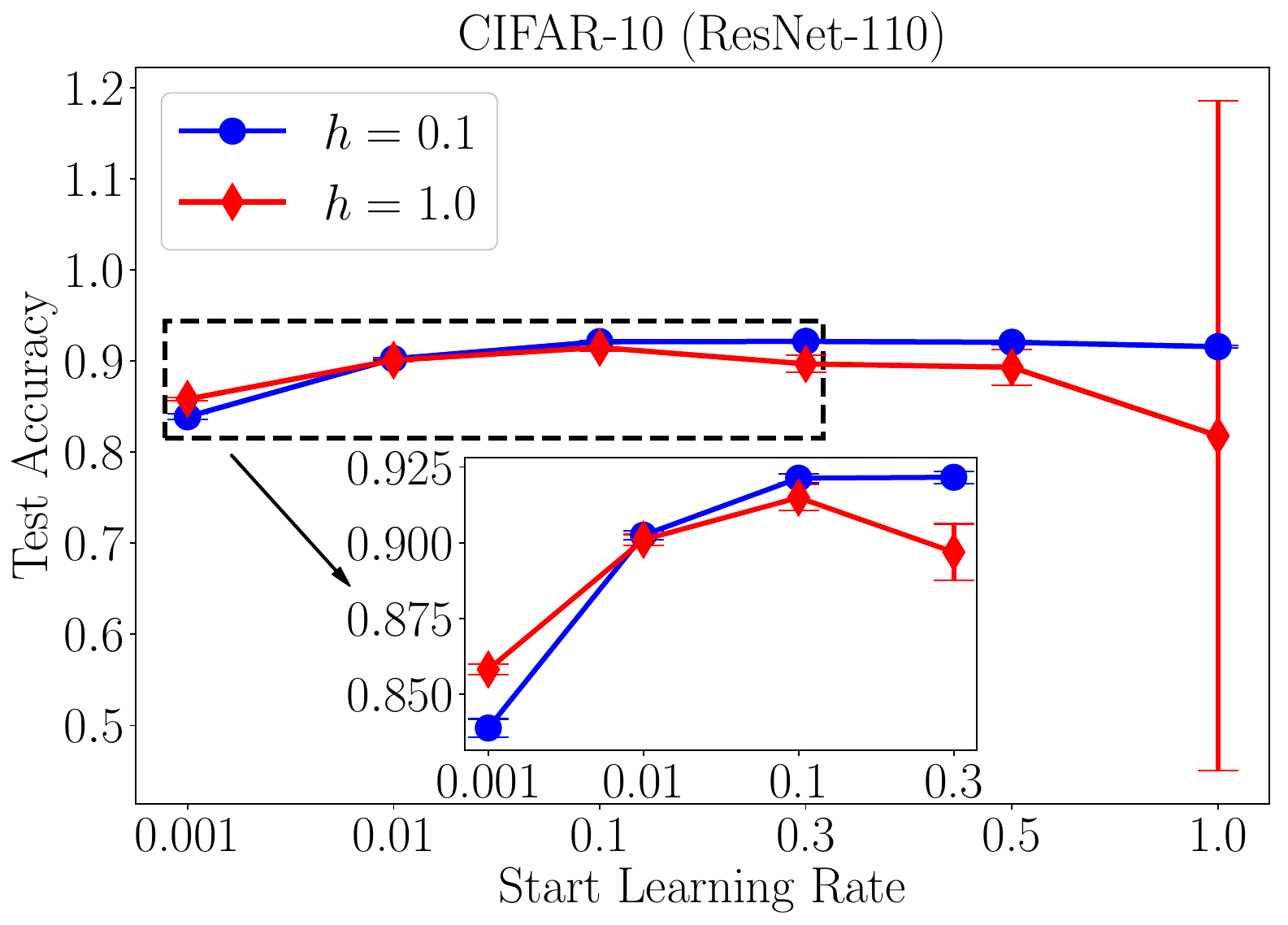}
	\includegraphics[scale=0.25]{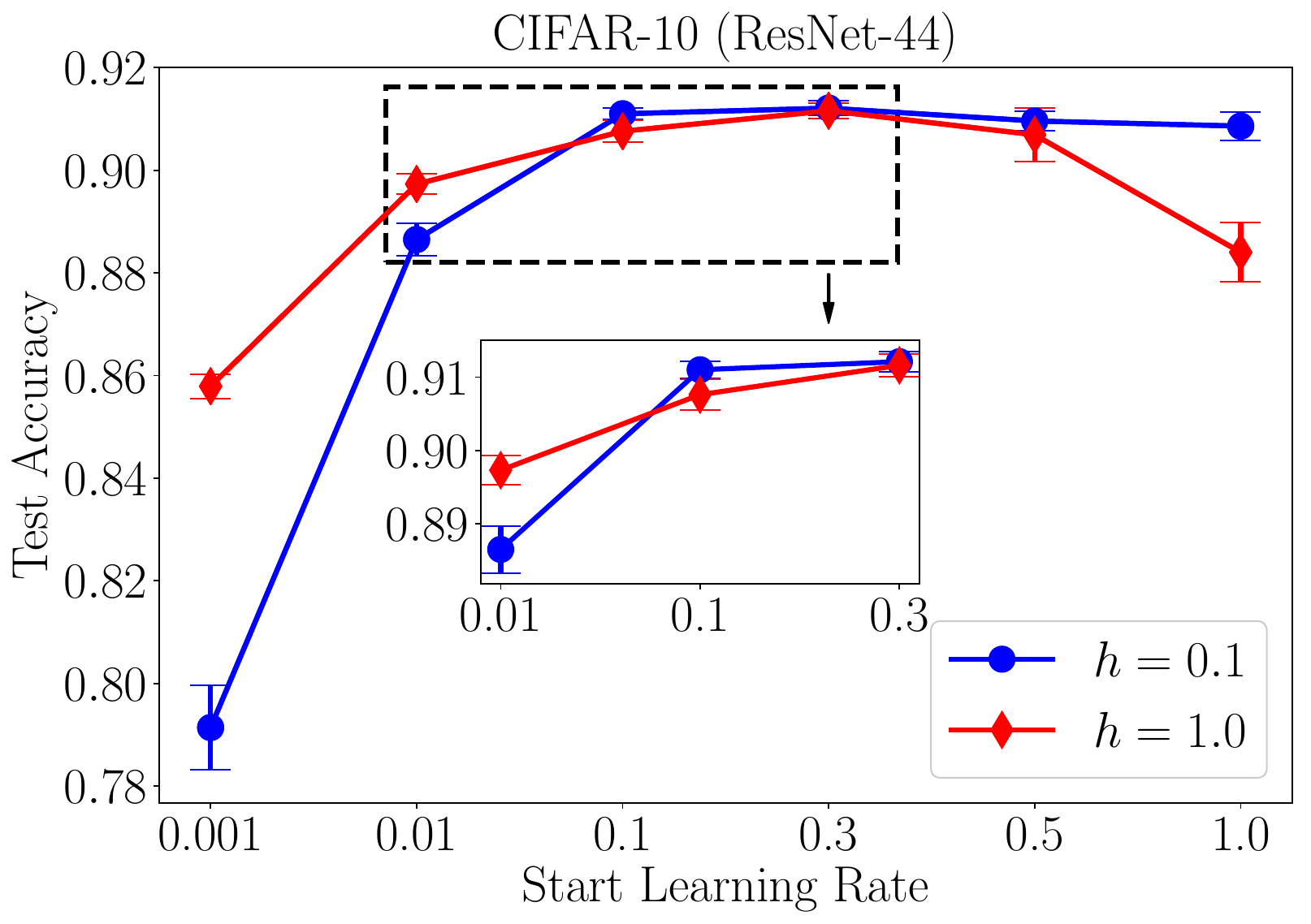}
	\caption{Training robustness comparisons of ResNet with different learning rates on {CIFAR-10}. 
	 }
	\label{fig:vis_different_leanring_rate_cifar_10}
\end{figure}

\begin{figure}[h!]\centering
	\includegraphics[scale=0.35]{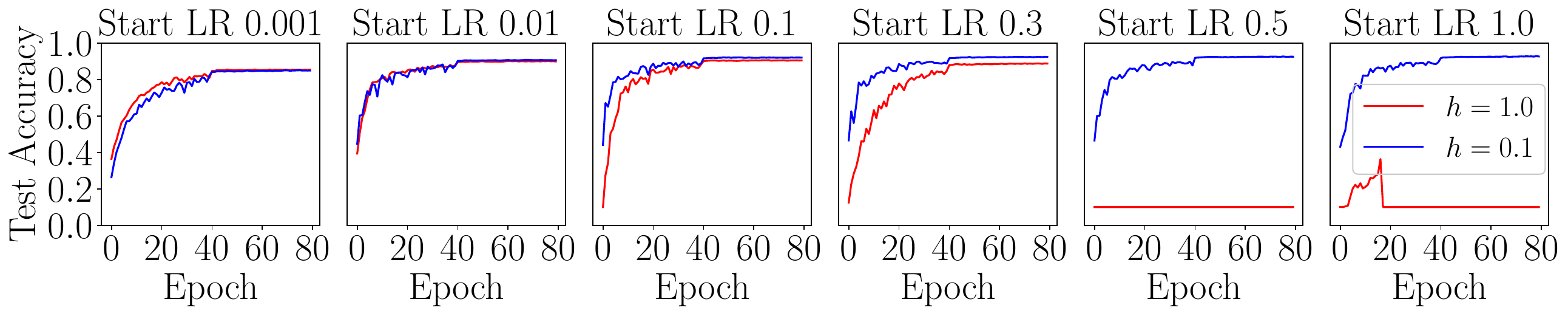}
	
	\caption{Visualization of convergence rate with different learning rates (LR). We train ResNet-218 on CIFAR-10, providing test accuracy over epochs.}
	\label{fig:vis_different_leanring_rate_convergence}
\end{figure}

\begin{figure}[h!]\centering
	\includegraphics[scale = 0.25]{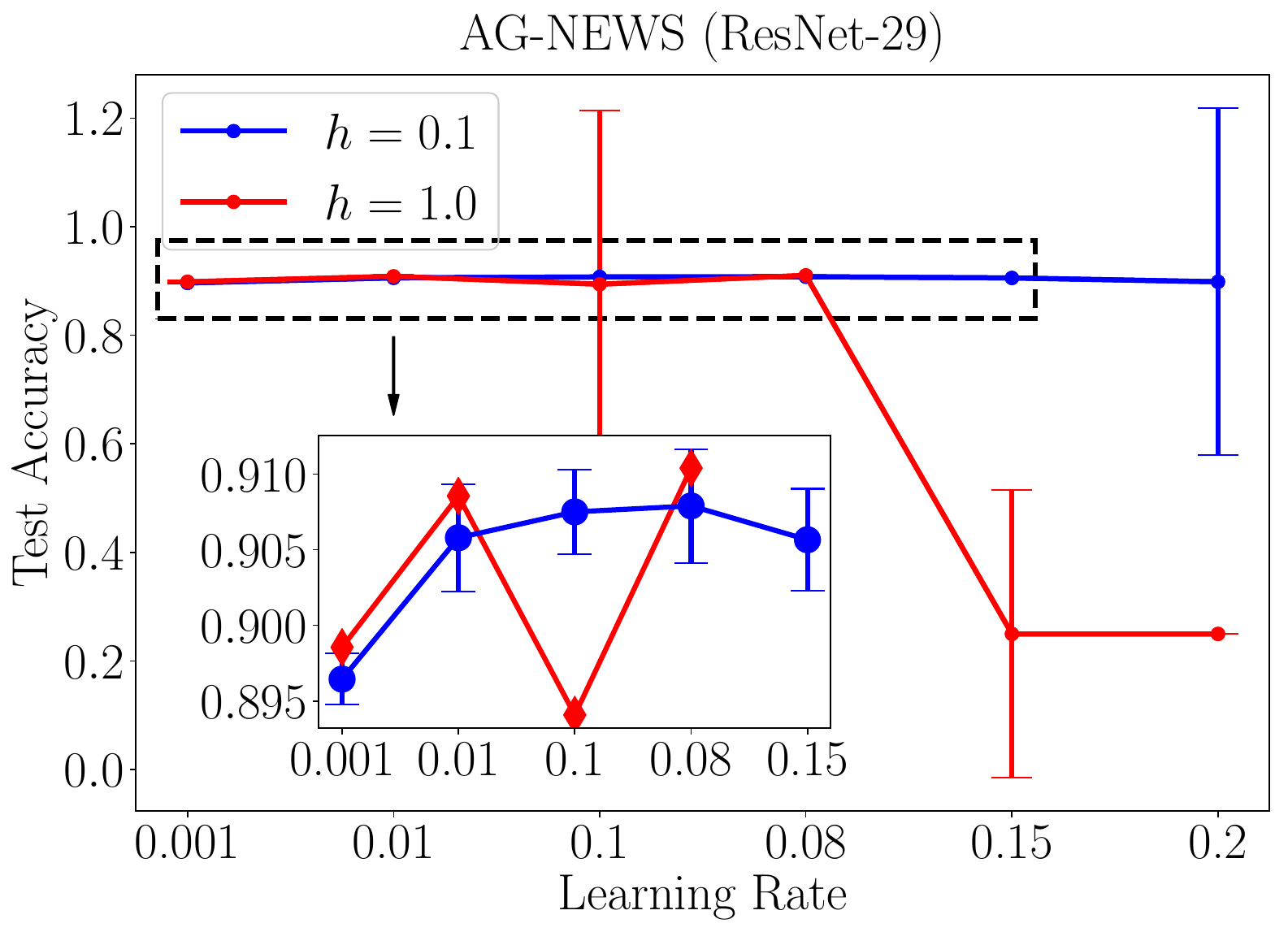}
	\includegraphics[scale = 0.25]{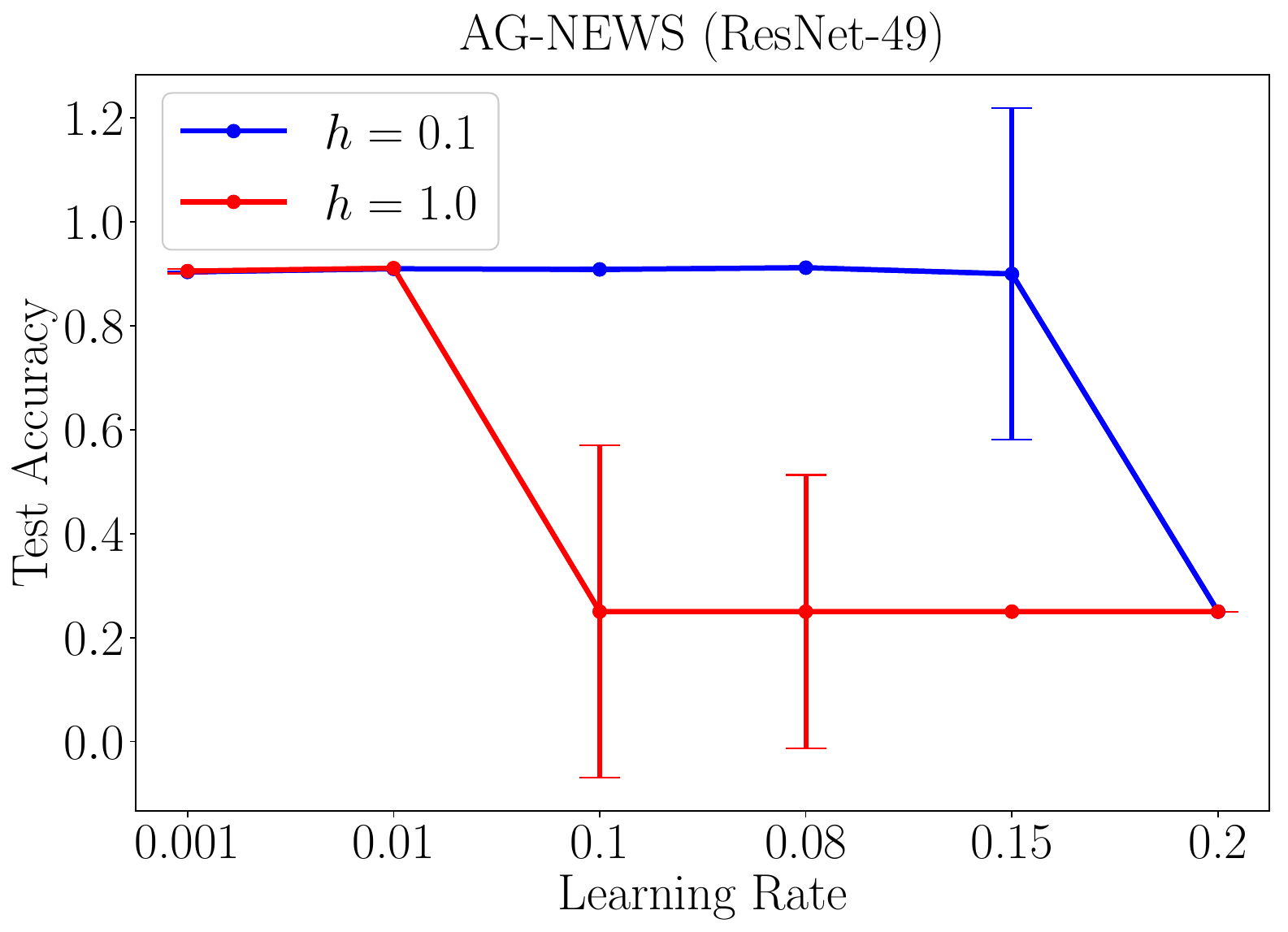}
	\caption{Training robustness comparisons of ResNet with different learning rates on {AG-NEWS}. 
	}
	\label{fig:vis_different_leanring_rate_ag_news}
\end{figure}

\clearpage

\section{Small $h$ Helps Networks On a Different Optimizer - ADAM}
\label{Appendix:small_h_on_adam}
\begin{figure}[h!]
	\centering
	\includegraphics[scale=0.25]{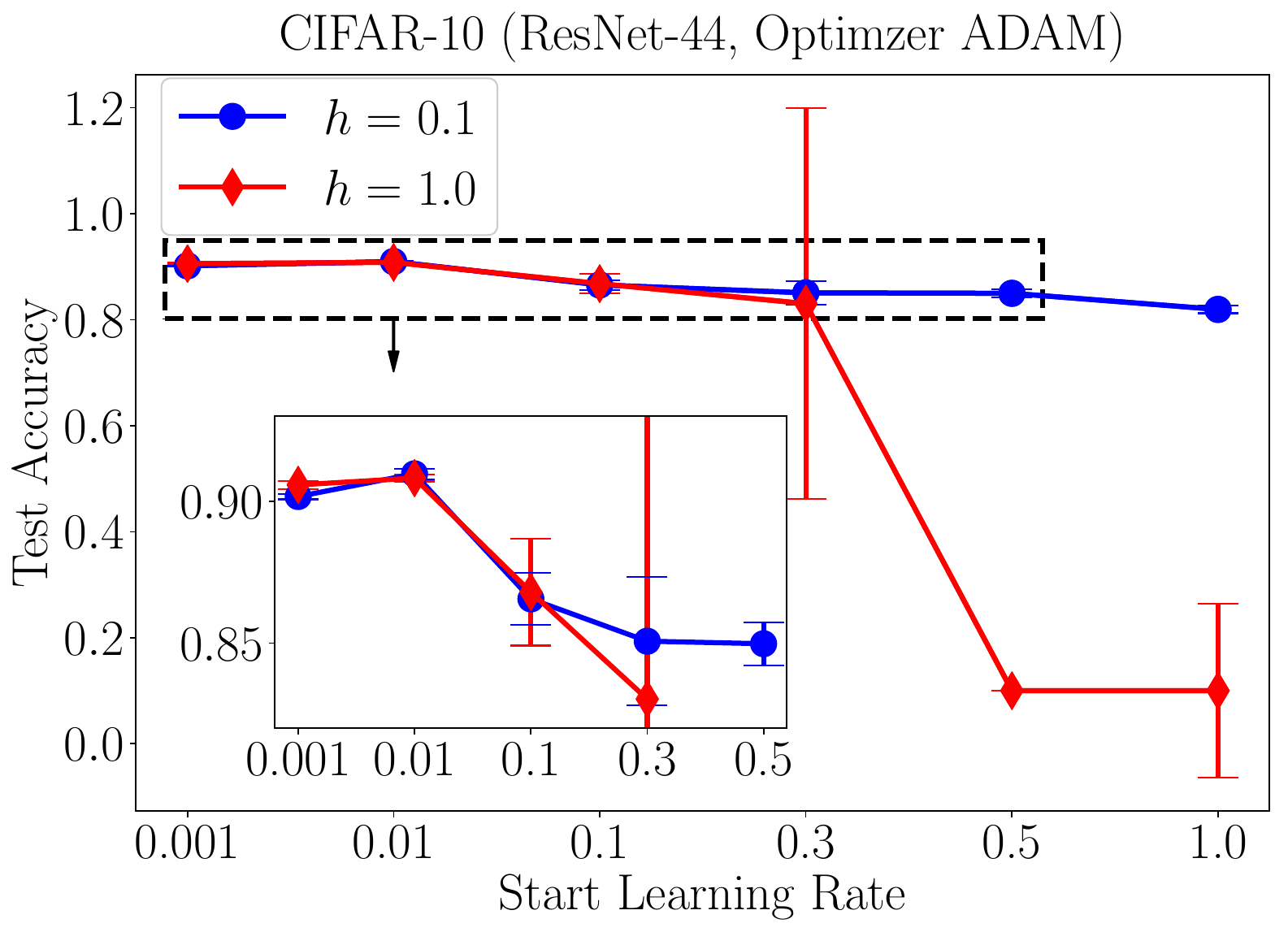}
	\includegraphics[scale=0.25]{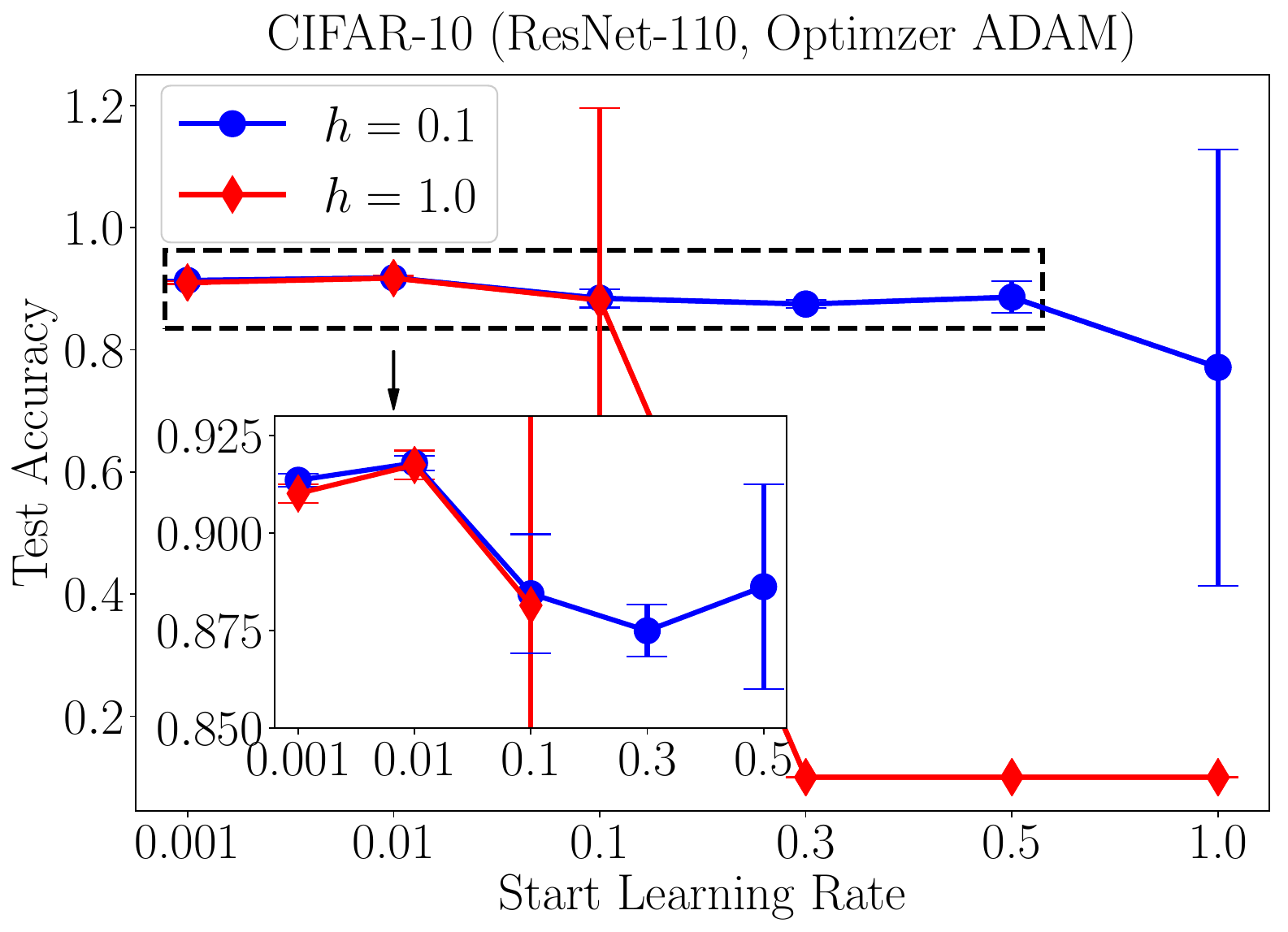}
	\caption{Training robustness comparisons of ResNet with different learning rates using ADAM optimizer. 
	}
	\label{fig:appendix_diferent_optimizer_adam}
\end{figure}

\section{Comparisons on Different Types of Weight Initialization} 
\label{Appendix:Comparison_diff_initialization}
We carry on experiments comparing different ways of weight initialization. Our experiments show that our method (small $h$) is still competitive in using different types of initialization to train ResNet with various depths.

\begin{figure}[h!]
	\centering
	\includegraphics[scale =0.23]{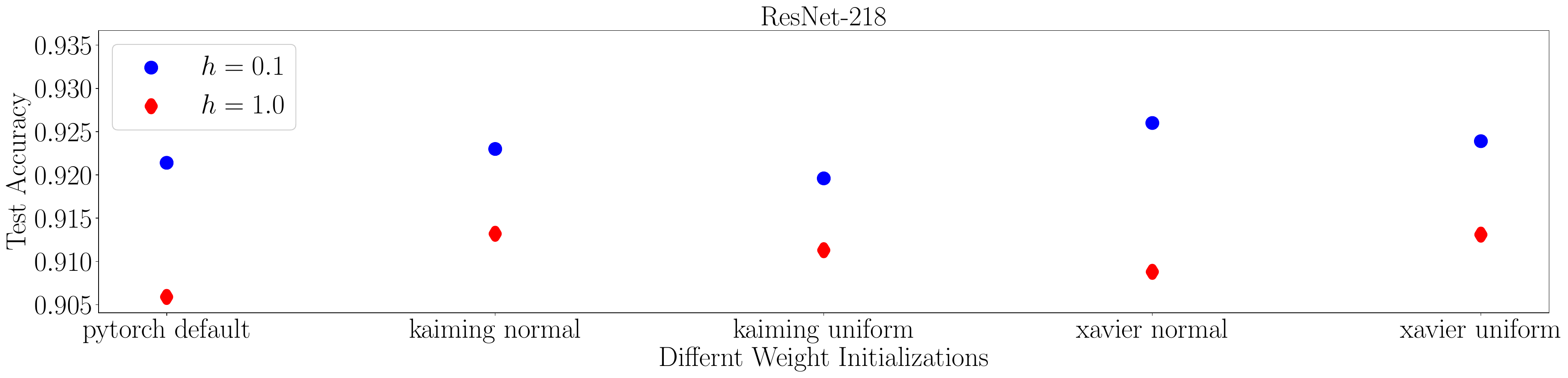}
	\includegraphics[scale =0.23]{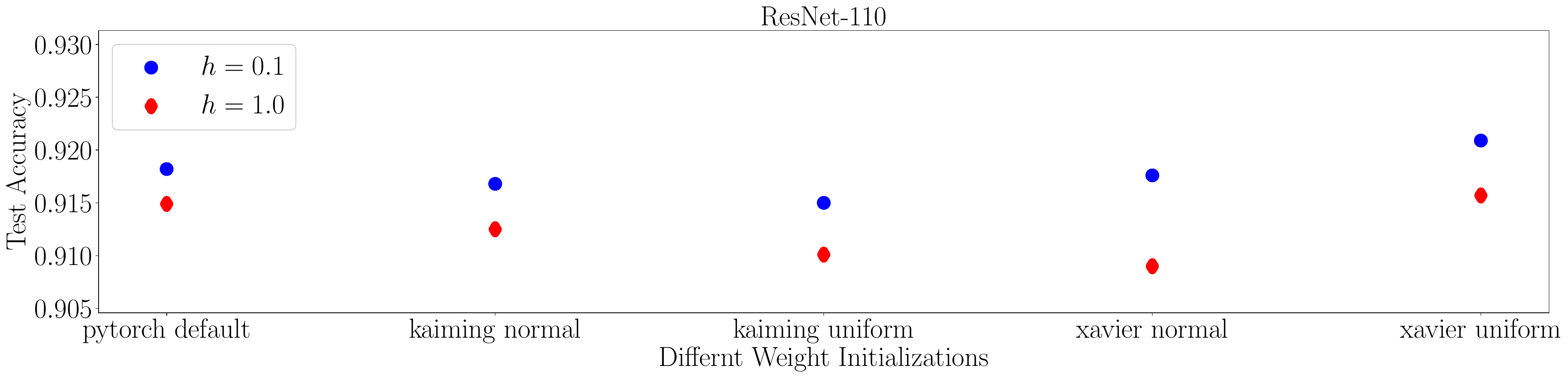}
	\includegraphics[scale =0.23]{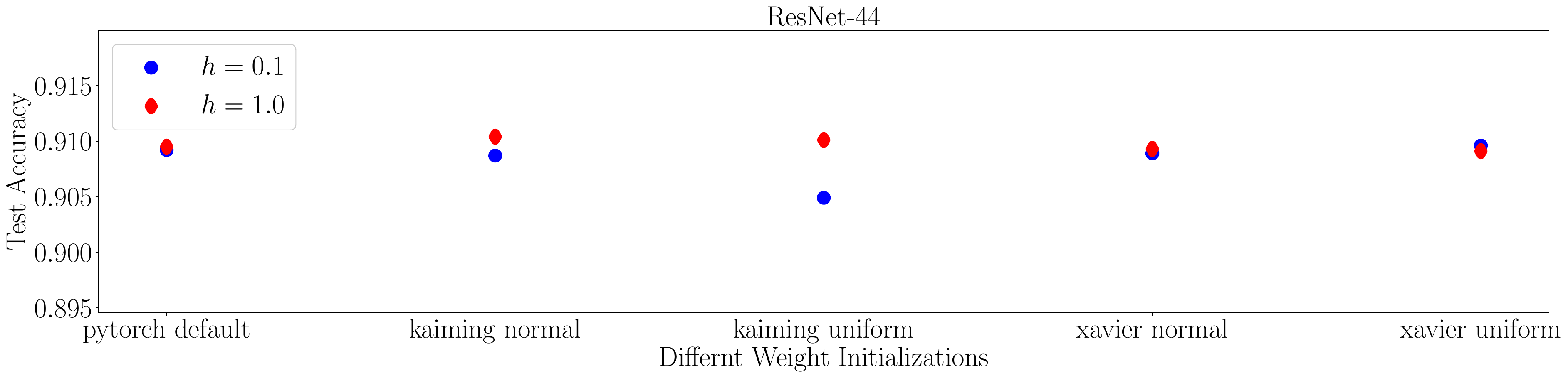}
	\caption{Comparisons on different types of weight initialization. We train ResNet on {CIFAR-10}, providing the test accuracy over different ways of weight initialization. }
	\label{fig:appendix_comparision_on_different_weight_init}
\end{figure}

\clearpage

\section{Visualizations on Noisy Input}
\label{Appendix:vis_noisy_input}
In Section~\ref{sec:generalization-robustness}, we train ResNet with $h = 0.1$ and $h = 1.0$ on noisy data (i.e., input has perturbations) and test it on clean data. 
For the training dataset {CIFAR-10}, we inject Gaussian noise at every normalized pixel with zero mean and different standard deviations (different noise levels). For the training dataset {AG-NEWS}, we randomly choose different proportions (different noise levels) of characters in the texts and alter them.
\begin{figure}[h!]
	\centering
	\includegraphics[scale = 0.5]{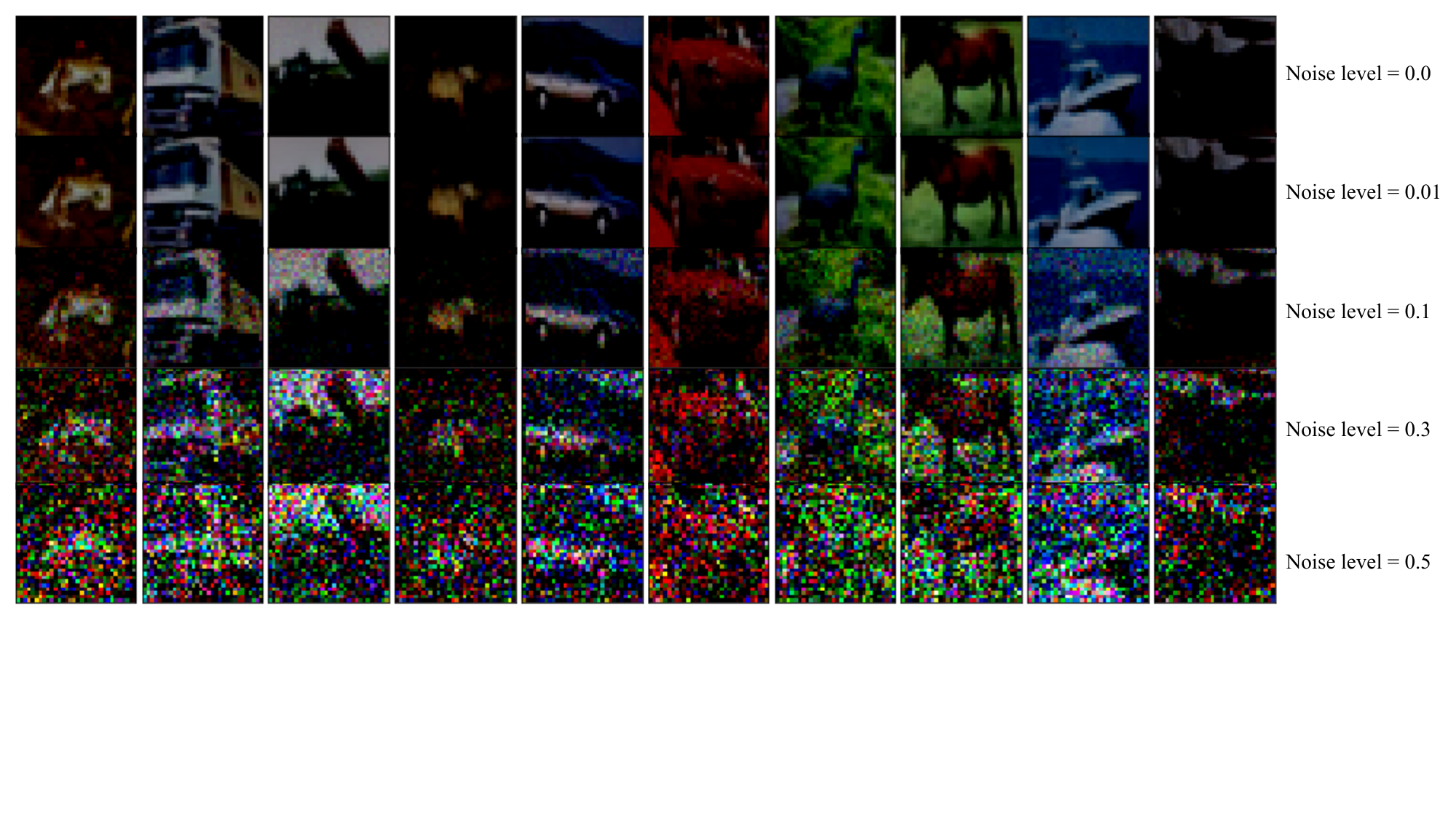} 
	\includegraphics[scale = 0.5]{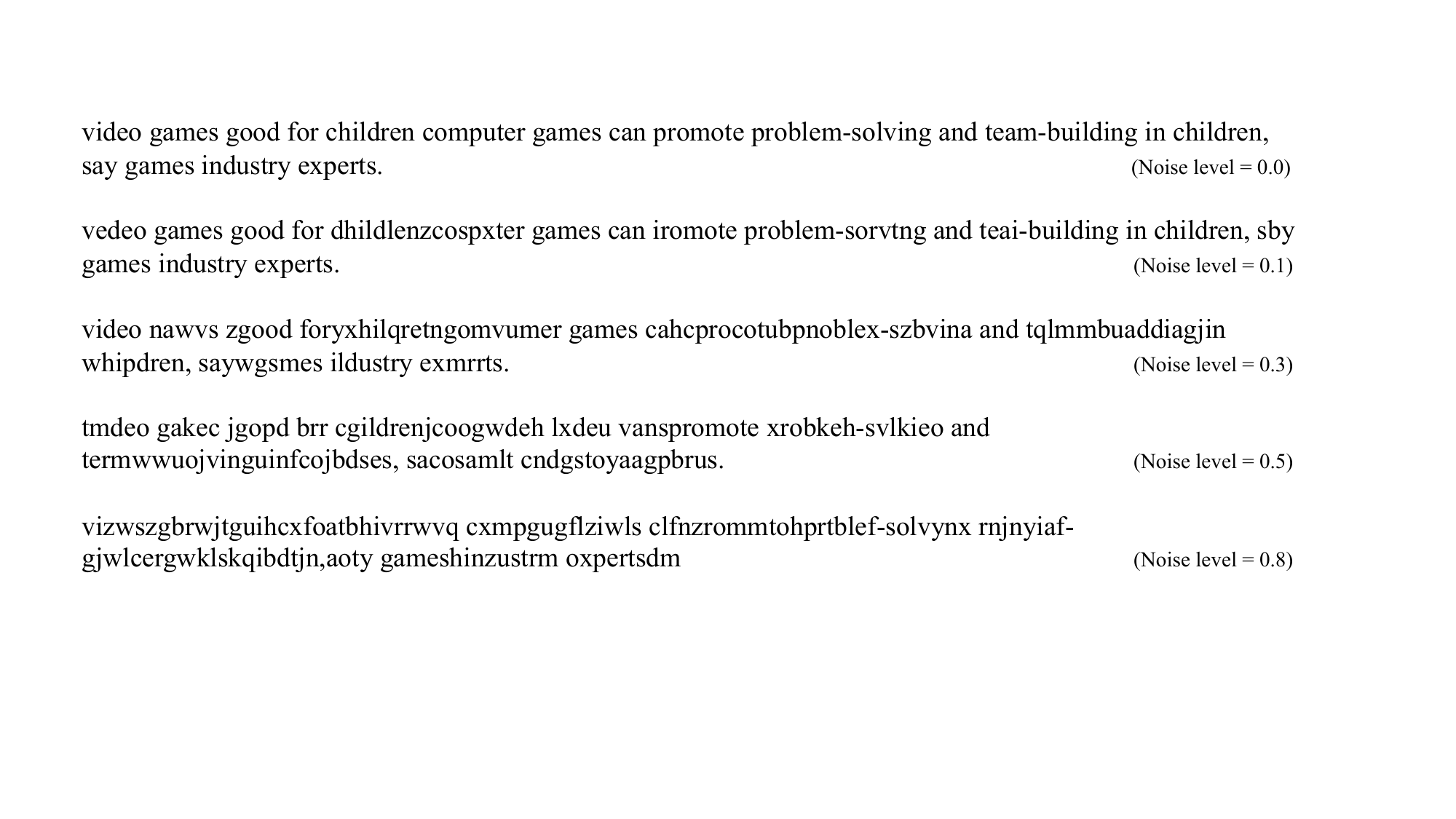}
	\caption{Noisy input visualizations of {CIFAR-10} and {AG-NEWS} }
\end{figure}

\end{document}